\documentclass[sigconf,nonacm]{acmart}
\AtBeginDocument{%
  }

\usepackage{enumitem}
\usepackage{amsthm}
\usepackage{makecell}  
\usepackage{algorithm}
\usepackage{algpseudocode}
\usepackage{graphicx}     
\usepackage{subcaption}   
\usepackage{caption}      
\usepackage{float}        
\usepackage{multirow}
\usepackage{booktabs}
\usepackage{array}
\usepackage{tikz}
\usetikzlibrary{trees,decorations.pathreplacing,calc}
\usepackage{caption}
\usepackage{subcaption}
\usepackage[capitalise,noabbrev]{cleveref}
\usepackage{balance}

\usepackage{soul}
\usepackage{xcolor}
\sethlcolor{cyan!15}

\usepackage{mdframed}
\newmdenv[
    backgroundcolor=gray!12,
    linecolor=gray!12,
    roundcorner=2pt,
    skipabove=4pt,
    skipbelow=4pt,
    innerleftmargin=6pt,
    innerrightmargin=6pt,
    innertopmargin=3pt,
    innerbottommargin=3pt,
]{findingshade}

\newcommand{\finding}[1]{%
\begin{findingshade}
\textit{\textbf{Summary:}~#1}
\end{findingshade}
}

\copyrightyear{2026}
\acmYear{2026}
\setcopyright{cc}
\setcctype{by}
\acmConference[CCS '26]{Proceedings of the 2026 ACM SIGSAC Conference on Computer and Communications Security}{November 15--19, 2026}{The Hague, Netherlands}
\acmBooktitle{Proceedings of the 2026 ACM SIGSAC Conference on Computer and Communications Security (CCS '26), November 15--19, 2026, The Hague, Netherlands}
\acmDOI{10.1145/3830454.3832707}
\acmISBN{979-8-4007-2871-6/2026/11}

\begin{document}

\title[Auditing Memorization in Mobility Prediction Models]%
{Secrets Everywhere: \\
Auditing Memorization in Mobility Prediction Models}

\titlenote{This manuscript is the full version of the paper accepted for
publication at the ACM SIGSAC Conference on Computer and Communications
Security (CCS 2026). It includes the complete appendices omitted from the
conference proceedings due to space constraints.}

\author{Anne Josiane Kouam}
\affiliation{%
  \institution{Inria \& TU Berlin}
  \city{Palaiseau}
  \country{France}
}

\author{Hristo Boyadzhiev}
\affiliation{%
  \institution{BIFOLD \& TU Berlin}
  \city{Berlin}
  \country{Germany}
}

\author{Konrad Rieck}
\affiliation{%
  \institution{BIFOLD \& TU Berlin}
  \city{Berlin}
  \country{Germany}
}

\begin{abstract}
    Human mobility prediction models, which forecast the next location in a user’s trajectory,
    are increasingly deployed in urban analytics, navigation, and personalized services. Yet, little is known about their potential to memorize and expose sensitive user trajectories from training data. 
    While memorization has been extensively studied in language models, mobility prediction poses unique challenges: training sequences encode human behavior at various spatial and temporal scales, creating privacy risks at different granularities.

    In this paper, we conduct the first systematic audit of memorization in mobility prediction models. While prior work has shown that privacy leaks can arise from such models, we systematically assess and quantify memorization risks at scale. We identify key challenges, including the lack of a randomness space, the multi-scale structure of trajectories, and user-specific behavioral diversity. 
    To address these challenges, we introduce a framework to quantify mobility memorization at different levels of granularity: individual locations, anchor pairs, and subtrajectory segments. We also develop user-grounded reference sets to assess how likely a model is to prefer training data over realistic alternatives.
    Our evaluation across multiple models and datasets reveals pervasive memorization patterns that correlate with user regularity and increase the risk of data extraction at inference time. Our findings call for mandatory privacy auditing in mobility prediction models.

\end{abstract}

\ccsdesc[500]{Information systems~Location based services}
\ccsdesc[500]{Security and privacy~Privacy-preserving protocols}
\ccsdesc[500]{Computing methodologies~Machine learning}


\keywords{Location Privacy; Spatiotemporal Data; Sequence Modeling;  Trajectory Extraction}

\received{20 February 2007}
\received[revised]{12 March 2009}
\received[accepted]{5 June 2009}

\maketitle

\section{Introduction}
\label{sec:intro}


Human mobility prediction plays a central role in modern urban analytics and decision-making, powering applications from smart transportation and crowd management to resource planning and personalized services. 
These models learn from historical location data to forecast individuals’ next positions, enabling governments to optimize public transit and reduce congestion~\cite{MobilityPolicyMaking}, while companies like Uber and mobile platforms anticipate demand and provide real-time, location-based recommendations~\cite{UberForecasting, dataplor2025_MobilityLocationData}.

This rise in practical applications has been mirrored by a growing body of research seeking to understand and model human mobility~\cite{survey1, survey2}. Early approaches relied on statistical formulations such as Markov chains or probabilistic transition matrices~\cite{survey3}, while more recent efforts have explored a range of sequential deep learning architectures, including RNNs, LSTMs, attention-based models~\cite{DeepMove, LSTPM, RNN1}, and combinations with convolutional approaches~\cite{graphflashback}. 
These models typically adapt architectures from language modeling to geospatial data, achieving notable performance with 
accuracies sometimes exceeding 90\%~\cite{Gebrie:2019}.

However, this predictive power might hide a privacy risk that has received little attention in the context of mobility prediction: \citet{carlini2019secretsharer} demonstrates that sequential models can unintentionally store and reproduce sequences from their training data; a behavior that persists even in the absence of classical overfitting.
This phenomenon, known as \textit{memorization}, implies that models can regurgitate specific training samples verbatim under the right inputs.
The risk is particularly acute for mobility sequences, where every data point corresponds to the real-world movement of an individual and thus carries intrinsic sensitivity.

We consider a realistic setting in which an adversary can interact with a trained
mobility prediction model through its standard prediction interface (e.g., by
querying location prefixes) and attempts to infer or reconstruct
sensitive portions of training trajectories. 
For instance, a recommender system trained on commuters’ GPS traces, when prompted with the known workplace of an individual, might continue the sequence with the exact locations that person typically visits after work. The returned data may include sensitive locations, such as home address or a medical clinic, as well as trajectory subsegments such as particular routes through the city.

In deep learning, concerns about unintended memorization are typically addressed through \textit{privacy auditing}, which systematically evaluates how much information from the training data can be inferred from a model, enabling practitioners to detect and mitigate potential data leakage during or after training.
One of the most influential frameworks in this area is \textit{exposure}~\cite{carlini2019secretsharer}, which 
has become a standard diagnostic for large language models, integrated into the TensorFlow Privacy library~\citep{tensorflowprivacy}.
Exposure quantifies memorization by testing how much a model assigns unusually high likelihood to a known “secret” seen during training.
It does so by comparing the model’s likelihood for that secret against many randomly generated unknown sequences drawn from the same distribution.
If the known secret ranks significantly higher than random candidates, the model is said to have memorized it.

In this paper, we establish that the \textit{exposure} metric fails to capture the privacy risks specific to mobility data, for two main reasons:

\begin{itemize}[leftmargin=*]
    \setlength{\itemsep}{3pt}
    \item \textbf{Outliers vs. inliers.}
    Exposure relies on the assumption that secrets are \emph{outliers}: sequences irrelevant to the learning task and unhelpful for improving accuracy.
    Assigning high confidence to such an outlier (e.g., a phone number) is clear evidence of memorization.
    This logic fits natural language models, where there is a clear boundary between what should and should not be learned.
    In mobility prediction, however, the situation is reversed.
    Every trajectory and location that improve predictive accuracy are also inherently sensitive: the data needed for generalization and the information that must stay private coincide.
    Here, secrets are \emph{inliers} 
    and high confidence may reflect the faithful reproduction of common, yet privacy-breaching, behavioral patterns.

    \item \textbf{Isolated vs. pervasive.}
    Exposure also assumes that secrets are \emph{isolated}: rare, atypical sequences whose presence in the training corpus is limited to a few scattered occurrences. 
    Under this assumption, memorization can be meaningfully summarized by a single global score capturing how the model retains such anomalies in general. 
    In mobility data, however, sensitivity is \emph{pervasive}: every trajectory corresponds to an actual individual’s movements and therefore constitutes a potential leak. 
    Consequently, memorization must take into account \emph{every} training trajectory, making leakage intrinsically data-dependent. 
    This shift requires analyzing a distribution of trajectory-level memorization signals, from which model-level summaries such as the maximum or tail percentiles can be derived.

\end{itemize}

As a result, existing exposure-based audits may substantially underestimate
privacy risks in deployed mobility prediction systems, leaving sensitive user
trajectories vulnerable to inference or extraction attacks. These fundamental
differences call for rethinking how memorization should be measured in mobility
prediction models. In this research, we address this gap by designing a framework
to assess memorization at the level of individual training trajectories, guided
by the following research questions:
\begin{enumerate}[leftmargin=*]
\setlength{\itemsep}{3pt}
\item 
\textbf{How can we quantify mobility memorization?}
We address this challenge by introducing a methodology that separates what a model \textit{should learn}—the underlying mobility patterns and transition structures—from what it \textit{should not memorize}, i.e., the exact coordinates and user-specific locations.
This helps identify distinct dimensions of memorization such as sequence-level, anchor pair-level, and localized recall.
To quantify these effects, we introduce three privacy metrics for individual trajectories: an adapted \textit{exposure}, the \textit{exposure preference}, and the \textit{exposure magnitude}. Together, these metrics offer different perspectives on how a model memorizes original patterns over a reference set of realistic alternatives.

\item 
\textbf{Which mobility patterns influence memorization?}
Based on this data-centric view, we explore how the statistical properties of individual trajectories correlate with memorization.
Through an extensive empirical evaluation on three public mobility datasets, ranging from 6,132 to 99,610 users and exhibiting distinct mobility dynamics, we find that memorization is far from marginal.
In some settings, up to 85\% of training trajectories show strong memorization signals.
Moreover, models systematically memorize more the trajectories of users with routinary and low-entropy movements, revealing a clear link between mobility behavior and privacy risk.

 \item \textbf{How does model design impact memorization?} To explore this, we reproduce and compare a representative set of predictive mobility models covering both statistical and deep learning approaches~\citep{hochreiter1997long,graphflashback,DeepMove,norris1998markov},
This diversity allowed us to assess memorization across architectures, differing in complexity, temporal modeling capacity, and representational depth.
We observe that while all models exhibit memorization, its extent and distribution vary with architectural design, suggesting that specific modeling choices can amplify memorization tendencies.
Most notably, we found no correlation between our trajectory-level memorization metrics and \citet{carlini2019secretsharer}'s model-level exposure, indicating that these measures capture fundamentally different dimensions of privacy risk.

\item \textbf{Does memorization translate into extractability?}
Finally, we assess whether highly memorized trajectories are also easier to recover from trained models. Using greedy extractability proxies that simulate an attacker repeatedly querying the model with plausible prefixes, we find a clear correlation: the higher a trajectory’s memorization scores, the fewer attempts are needed to reconstruct it. This suggests that memorization not only reflects internal model behavior but also predicts real extraction risk at inference time.
\end{enumerate}


Taken together, our work thus makes three main contributions. 
(1)~We introduce a methodology for identifying unintended memorization in mobility prediction models.
(2)~We define complementary trajectory-level metrics that capture distinct memorization behaviors and enable privacy risk assessment. 
(3)~We conduct a systematic empirical study showing that memorization varies across trajectories, datasets, and model architectures, with direct implications for privacy auditing and extraction risk. 
Our findings indicate that memorization in mobility prediction models is widespread, underscoring the need for systematic privacy auditing before deployment in real-world mobility services.

\paragraph{Roadmap.}
We begin in \cref{sec:background} with background and related work, followed in \cref{sec:challenges} by the main challenges of memorization in mobility data. \cref{sec:methodology} presents our framework and metrics, and \cref{sec:experimental_setup} describes the experimental setup. Our empirical results are presented in \cref{sec:empirical_findings}, with broader implications discussed in \cref{sec:discussion}. We conclude in \cref{sec:conclusion}.

\paragraph{Reproducibility.}
To foster reproducibility, we release all code and evaluation scripts in an anonymous repository for the review process. Details are provided in the Open Science appendix (\cref{app:open_science}).

\section{Preliminaries}
\label{sec:background}

We start by outlining the foundations of human mobility prediction and then introduce the exposure framework, a widely adopted tool for quantifying memorization in language models.

\subsection{Mobility Prediction Models}

A human mobility trajectory is represented as an ordered sequence of locations
\[
T_u = (l_1, l_2, \ldots, l_n),
\]
where each \(l_i\) denotes the latitude–longitude coordinate visited by user \(u\) at the \(i\)-th time interval. 

In most datasets, time is discretized into regular slots such as 30 minutes or one hour, making the temporal ordering implicit in the index $i$ and allowing trajectories to be treated as fixed-grid sequences of numerical coordinates. We focus on this representation, although our approach can also be extended to trajectories with non-uniform time steps.


The goal of mobility prediction modeling is to forecast the next location an individual will visit, given their historical mobility data.
Formally, this entails learning a function
\[f_\theta: (l_1, \ldots, l_k) \longmapsto l_{k+1},\]
that maps a prefix of the trajectory to a predicted next location $l_{k+1}$ in the time grid. 
This sequential prediction task is analogous to language modeling, treating locations like words and trajectories like sentences, such that mobility patterns are modeled probabilistically in temporal order.

Over the past decade, a wide spectrum of model architectures has been explored for next-location prediction~\cite{survey1,survey2, survey3}.
Early approaches relied on \textit{Markov chains} and other probabilistic transition models to capture short-range dependencies between locations~\cite{markov1, markov2, markov3}. While effective with limited data, these traditional methods struggled to capture long-range temporal patterns. 
The advent of deep learning brought recurrent neural networks (RNNs) and their gated variants (LSTMs, GRUs), which enabled learning longer-term dependencies from raw trajectories~\cite{RNN1, RNN2}.
Subsequent advances introduced attention mechanisms and spatio-temporal embedding techniques to better model periodicity, context, and user-specific preferences.

As an example, the \textit{DeepMove} model augments a GRU-based sequence predictor with an attention layer to capture both long-term periodic behaviors and short-term sequential patterns~\cite{DeepMove}.
Similarly, \textit{LSTPM} model employs a context-aware network (with geo-dilated LSTM components) to integrate users’ long-term and short-term location preferences~\cite{LSTPM}.
Other innovations include models like 
\textit{Graph-Flashback}, incorporating graph neural networks over a spatial-temporal graph~\cite{graphflashback}. Recently, large language models (LLMs) have also been applied to mobility prediction in a zero-shot or few-shot setting~\cite{mobpred-LLM1, mobpred-LLM2}. However, since these models are not fine-tuned on target mobility datasets, evaluating their memorization remains out of this study’s scope.

Overall, deep sequential models have achieved strong performance in next-location prediction, often exceeding 80--90\% accuracy in urban datasets~\cite{survey2}.
For broader coverage of statistical and learning approaches, we refer the reader to recent surveys~\cite{survey1,survey2}.

\subsection{Memorization and Exposure}

In language models, \textit{memorization} refers to the model unintentionally preserving specific training examples that are irrelevant to its general task~\cite{carlini2019secretsharer}. \citet{carlini2019secretsharer} formalizes a way to quantify such memorization with an \textit{exposure} metric. They insert known unique sequences called \textit{canaries} into the training data (e.g.,~a random ID number within a sentence) and then measure how likely the trained model is to produce that exact sequence compared to all other possible sequences of the same format. This likelihood is measured via the log-perplexity, a proxy for the model’s confidence in generating a given string. The set of all possible canary values defines a \textit{randomness space} \(R\). 

By ranking all candidate canaries by their model likelihood, the exposure of a specific secret canary is defined as:
\begin{equation}
\label{eq:exposure}
\operatorname{exposure}(s)
= \log_2 |R| \;-\; \log_2\!\big( \operatorname{rank} \left (L(s) \right ) \big)
\end{equation}
where \(rank(L(s))\) is the position of the true secret in the model’s likelihood ranking. Intuitively, this value represents how much the model reduces the uncertainty of guessing the secret. An exposure of 0 means no memorization, while a higher exposure indicates that the model has significantly memorized the canary.

The study by \citet{carlini2021extracting} has prompted the development of different defenses. For example, deduplicating text corpora can reduce memorized outputs~\citep{lee2022deduplication}, while post-hoc approaches such as machine unlearning aim to remove memorized examples without retraining~\citep{unlearning2023rethinking}. Similarly, the exposure metric has also influenced industry practices, including audits of production systems and the adoption of privacy-preserving tools like the TensorFlow Privacy library \citep{tensorflowprivacy}.
Despite this impact, research on memorization has remained concentrated on natural language \citep{memorization_survey} and, more recently, on image generation models \citep{memorizationdiffusionmodels, solidmarkevaluatingimagememorization}. Comparable evaluations in other domains, particularly mobility data, are still scarce, which motivates the design of our auditing framework.

\subsection{Privacy in Mobility Prediction Models}

Mobility prediction models, especially those based on deep learning, have been recognized as potential vectors of privacy leakage. 
While raw and pseudonymized mobility traces have long been known to be highly identifying~\cite{deMontjoye2013}, only recently has the literature begun to uncover that predictive models trained on such data can also encode and leak sensitive user information. 

Several studies have demonstrated various forms of leakage: 
model inversion attacks can reconstruct parts of users' trajectories from personalized predictors~\cite{atrey:2021}, while membership inference attacks 
can determine whether a user's data or a specific location was used in training, either at the individual~\cite{Cai:2024} or aggregate level~\cite{Pyrgelis:2018}. 
These findings reveal that predictive models can become as privacy-sensitive as the mobility data they are trained on. This realization is especially critical given the growing deployment of such models in real-world systems, where they can be probed or exploited without directly accessing raw data.

\paragraph{Positioning.} 
In this paper, we build on existing work on privacy risks in mobility prediction models but go beyond showing that (and how) leakage can occur. Instead, we introduce the first framework for providing a quantitative and thus measurable account of this leakage. As inference-time leakage stems from models internalizing sensitive training data, memorization provides the ideal signal through which we can examine this behaviour.
By quantifying how much user-specific information is preserved, memorization offers a general-purpose diagnostic tool useful both for anticipating leakage and for evaluating the effectiveness of privacy-preserving techniques, establishing a foundation for privacy auditing in the domain of mobility prediction.

\section{Challenges in Mobility Memorization}
\label{sec:challenges}

Diagnosing memorization in mobility models poses several unique challenges due to the structure of the data, the user-specific nature of mobility, and the complexity of model behavior. 
Unlike prior work in text domains, where canaries can be synthetically generated and inserted at scale, mobility traces require
careful design of user-grounded references 
that preserve realism, structure, and spatial semantics. 
We outline the key conceptual and technical obstacles that shaped our auditing framework.

\subsection{Multiple Scales of Memorization}
\label{subsec:challenge1}



Mobility trajectories are not flat sequences; they are structured behaviors composed of recurring locations, characteristic transitions, and sub-trajectories that reflect personal routines. 
This means that memorization can arise at several levels: a model may not need to retain an entire trajectory to compromise privacy; it may instead retain a specific location, an association between two points, or a movement pattern (e.g., home $\rightarrow$ gym $\rightarrow$ café). As a consequence, sensitive information can occur at any scale, from highly local to globally scattered across the trajectory.

This multi-layered structure is not fully captured by exposure, which tracks only one form of memorization: whether a model assigns disproportionately high likelihood to an \textit{exact} sequence or token compared to plausible alternatives. 
In mobility data, however, privacy leakage can also stem from information beyond exact sequences that may still be sufficient to identify a user.
This demands an assessment of how different patterns within a trajectory are retained and reemerge in the model’s behavior.

\subsection{Complex Randomness Space}
\label{subsec:challenge2}

Exposure quantifies memorization by comparing a known training sequence against many left-out candidates drawn from a randomness space, typically random strings that are meaningless to the model. This works for language models, where secrets are rare, context-independent outliers (e.g., ID numbers). 
In mobility prediction, however, we are concerned with the memorization of inlier sequences, i.e., realistic user trajectories that the model must learn from, yet should not regurgitate. Hence, the equivalent of a randomness space becomes much harder to define. Randomly sampling locations produces implausible trajectories that the model easily rejects by giving lower likelihood, not because it memorizes, but because the comparison set is behaviorally incoherent.

To measure whether a model memorizes a user's trajectory, we thus need reference sequences that are both \textit{unseen} and \textit{plausible}, capturing realistic patterns like commuting rhythms, temporal regularity, and geographic coherence. Moreover, because mobility behavior varies from person to person, these references must be constructed individually to reflect each user’s movement style without replicating their true path. 

\subsection{User-Centric Evaluation}
\label{subsec:challenge3}

Finally, in mobility data, each training example corresponds to a specific individual, making memorization inherently user-specific. A model that stores part of one user’s routine poses a privacy risk to that individual, even if it generalizes well overall.
This user-centered nature of mobility brings added complexity: the number of plausible alternatives against which memorization should be assessed is not the same for every user. 
Differences in mobility entropy, visited locations, and routine imply that the randomness space $R$ used in exposure (Eq.~\ref{eq:exposure}) varies across users. 


Hence, exposure values are not directly comparable in the context of mobility: 
a high rank (strong memorization) with a small $|R|$ yields lower exposure than the same rank in a large reference set.
Evaluating memorization, therefore, requires metrics that remain comparable despite heterogeneous random spaces, enabling a fair assessment of how much the model preserves data.

\section{Methodology}
\label{sec:methodology}

\begin{figure}[h]
    \centering
    \includegraphics[width=\linewidth]{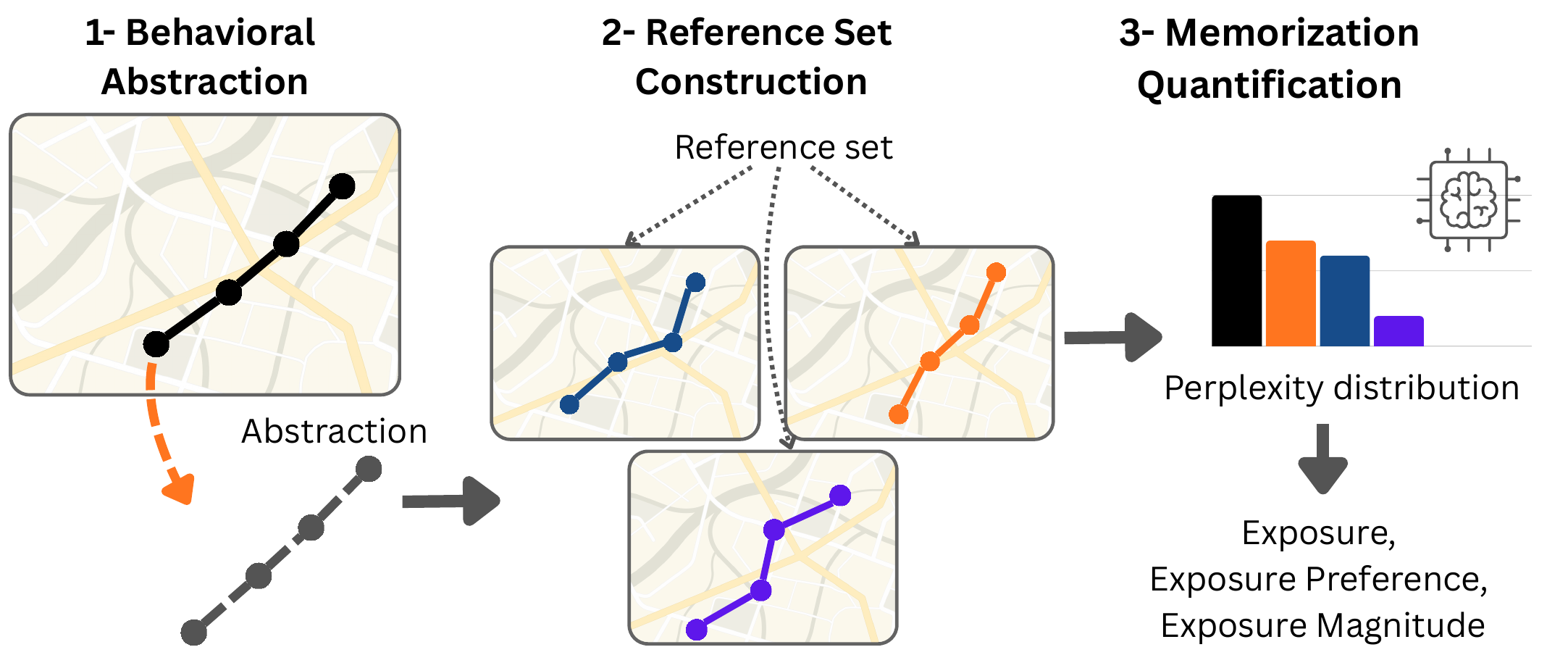}
    \caption{Overview of our memorization auditing framework.}
    \label{fig:framework-overview}
\end{figure}

To address the challenges in mobility memorization, we present a framework for quantifying how much a mobility prediction model memorizes individual training trajectories. The core idea is to distinguish what the model is expected to learn from what it could but should not memorize. Our evaluation proceeds in three conceptual steps, illustrated in \cref{fig:framework-overview}.

\begin{enumerate}[leftmargin=*]
    \setlength{\itemsep}{4pt}
    \item \textbf{Behavioral abstraction:}
    For a given training trajectory, we define an abstraction that captures the structural pattern the model is expected to generalize from, such as mobility routines, temporal rhythms, or spatial transition types, while deliberately discarding user-specific identifiers
    ($\rightarrow$ \cref{subsec:abstraction}). 
    
    \item \textbf{Reference set construction:} From the abstraction, we generate a reference dataset that reflects the distribution of plausible, behaviorally consistent alternatives. These reference trajectories are sampled from other users who exhibit similar patterns, ensuring they share high-level mobility structure but differ in sensitive content
    ($\rightarrow$ \cref{subsec:reference}).
    
    \item \textbf{Memorization quantification:}
    We measure how much more likely the model is to assign high confidence to the target trajectory compared to its reference set. Intuitively, a non-memorizing model should treat the original and reference trajectories similarly. A strong preference for the original trajectory indicates memorization
    ($\rightarrow$ \cref{subsec:quantify}).

\end{enumerate}

\subsection{Behavioral Abstraction}
\label{subsec:abstraction}

Mobility trajectories encode diverse facets of human behavior in the same data: a single sequence simultaneously reflects a user’s global movement style (e.g., distance range, regularity), anchor points such as home and work, 
and fine-grained routines embedded within specific time windows. 
This hierarchical structure makes it difficult to define a single notion of memorization (see \cref{subsec:challenge1}). Indeed, a model may successfully generalize across some behavioral aspects, such as displacement rhythms, while memorizing others, such as 
locally repeated sub-sequences. 


To address this, we express memorization in mobility prediction as a spectrum of risks rather than a binary property. 
Precisely, we distinguish several \textit{memorization risks}, each targeting a specific type of behavioral information conveyed by the trajectory. For each risk, we specify what the model is expected to generalize versus what would constitute memorization. This decomposition allows us to isolate how models behave across different behavioral layers.
To formalize each risk, we define a corresponding \textit{trajectory abstraction} as a simplified representation of the signal we expect the model to generalize from, not memorize. 
\begin{definition}[Trajectory Abstraction]
Let $T_u$ denote a discretized trajectory of user $u$. 
A \emph{trajectory abstraction} is a mapping
\[
A: T_u \mapsto Z_u \in \mathbb{R}^d
\]
that projects the raw trajectory into a fixed-size representation $Z_u$ capturing the mobility patterns that the model is expected to generalize from rather than memorize.
\end{definition}

Based on this abstraction, the memorization risk can be measured by how strongly the model distinguishes the specific training trajectory $T_u$ from other sequences $T'_u$ that satisfy the same abstraction, that is, $A(T'_u) = A(T_u) = Z_u$. The particular definition of $A$ allows us to encode different types of memorization and quantify their presence in a model. In the following, we introduce three memorization risks and corresponding definitions of $A$, spanning from the retention of locations to leakage of movement routines. Although not exhaustive, these provide a  structured lens to assess memorization across mobility properties.


\subsubsection{Location memorization}
\label{subsubsec:location}
This memorization risk arises when a model learns the specific locations visited by a user, rather than generalizing from the behavioral mobility patterns.
For example, consider a user who remains stationary all day while teleworking. The desired behavior from the model is to learn that full-day stationarity is realistic in residential areas, not to memorize the specific coordinates of the user’s home. Likewise, when people commute from home to work in the morning, the model should learn the pattern of commuting from residential to business zones, not the precise route or destinations of an individual.

From the perspective of abstraction, the model should thus generalize over \emph{how} people move rather than \emph{where}. The relevant pattern here is the user's displacement dynamics, expressed as the sequence of motion vectors over time, independent of absolute location.
Let $T_u = (l_1, \ldots, l_n)$ denote the trajectory of user $u$, where $l_i$ is the location visited at time step $i$.
Consequently, we define the abstraction $A$ as a sequence of first-order displacements:
\[
A_{\text{motion}}(T_u) = ( \Delta l_2, \dots, \Delta l_n ) \text{~~with~~} \Delta l_i = l_i - l_{i-1}.
\]
This abstraction removes absolute positioning, retaining only the user's trajectory shape. It captures whether the user remains stationary, their typical movement range, and directional tendencies, while eliminating specific locations that could be memorized. Each time step is treated uniformly, ensuring that mobility flow is preserved but spatial identity is removed. In other words, we can measure memorization in a model when it associates concrete locations with a trajectory $T_u$, even though learning $A(T_u)$ would have been sufficient for generalization.

\subsubsection{Anchor-pair memorization}
\label{subsubsec:anchor}
This risk occurs when a model memorizes the \emph{association} between salient anchor locations frequently visited by a user, typically their home and workplace. Unlike the previous risk, which already captures the leakage of individual locations, here we specifically assess whether the model learns to link locations together in a way that can reveal the user's routines. Such anchor pairings are particularly sensitive because they form a behavioral signature: once one location is known, the model may trivially infer the other, even if not memorized independently. 
In this setting, however, we do not aim to prevent the model from learning that users exhibit anchor-based routines but we do expect it to generalize across users with similar patterns, rather than memorizing specific pairs. 


To capture this expected behavior, we represent anchor usage through a coarse abstraction: the location of the primary anchor such as home or work and the commuting radius $d$ associated with it. In the home-based variant, we define
\[
    A_{\text{home}}(T_u) = (l_{\text{home}}, d),
    \quad
    A_{\text{work}}(T_u) = (l_{\text{work}}, d),
\]
where $l_{\text{home}} \in \mathbb{R}^2$ is the centroid of early morning locations, such as those before 6am, and $d \in \mathbb{R}_+$ is the average distance to the user’s daytime activity region. Similarly, $l_{\text{work}}$ denotes the centroid of mid-day locations, for example, between 10am and 6pm.
%
%
While we focus on the two most common anchors, namely home and workplace, the approach also applies to any salient location that structures daily mobility, such as transport hubs, dinner spots, or activity centers.


\subsubsection{Segment-level Memorization}
\label{subsubsec:segment}
The final memorization risk targets localized patterns within a user’s trajectory, that is, short fragments that may reflect short routines or temporally localized behaviors. Unlike memorization of locations or anchor pairs, segment-level memorization concerns the model's tendency to rigidly learn specific short-term movement traces, even when they are not semantically essential.
For instance, a user may often visit different places in a particular order around lunchtime. While such a loop is behaviorally plausible, the model should not memorize the exact sequence of locations if the underlying routine can be expressed more flexibly. Memorization here would mean that the model fails to generalize across slight variations of this local pattern, treating one specific realization as canonical.


To capture this risk, we define the abstraction over a localized segment of the trajectory treated as an interchangeable placeholder. 
Specifically, for a continuous subsequence $T_u^{[s:e]} = (l_s, \dots, l_e)$, we abstract it as a masked slot within the full sequence: 
\[
A_{\text{segment}}(T_u) = (l_1, \dots, l_{s-1}, \star, l_{e+1}, \dots, l_n),
\]
where $\star$ denotes the masked segment that can be substituted by any plausible sub-trajectory.

This abstraction reflects the intuition that users may vary their short-term routines (e.g., different lunch routes or minor errand sequences) while preserving overall behavior. The goal is to ensure the model generalizes across such variations, without retaining one specific realization of a sub-pattern. 

\subsection{Reference Set Construction}
\label{subsec:reference}

An ideal model without memorization should generalize beyond specific trajectories to capture the underlying behavioral patterns rather than memorize where or when they occurred. For each trajectory $T_u$ used in training, we seek to define a reference set $\mathcal{R}(T_u)$ of unseen trajectories that express the same behavioral pattern, such that a model generalizing correctly should assign them comparable likelihoods. However, the construction of this set critically depends on how it abstracts from the particular target, and so we introduce an ideal and an approximate construction strategy.

\subsubsection{Ideal reference construction.} In the ideal case, we are able to synthesize new trajectories for the reference set that are (almost) indistinguishable from real behavior and originate from the same data distribution as the target trajectory.

\begin{definition}[Ideal Reference Set]
Given a behavioral abstraction~$A$ and a training trajectory $T_u$, the ideal reference set is the collection of held-out trajectories (i.e., not seen during training) that satisfy the same abstraction as $T_u$:
\[
\mathcal{R}(T_u) = \left\{ T'_v \in \mathcal{T}_{\text{held-out}} \mid A(T'_v) = A(T_u) \right\}.
\]
\end{definition}

Constructing such ideal reference sets, however, is only possible in very specific cases, as their feasibility depends strongly on the abstraction $A$ in question. While it might be possible to synthesize a trajectory $T'_v$ satisfying $A(T'_v) = A(T_u)$, we have limited control over whether it is truely realistic. For example, a commute between two residential areas may be plausible in one neighborhood but infeasible elsewhere because of barriers.

Fortunately, for \emph{segment-level memorization} (\cref{subsubsec:segment}), constructing realistic alternatives is feasible. We can replace the segment with a plausible alternative from the same user. 
In particular, we replace masked subsequences with alternatives drawn from the same user, including: (i) \textit{substitute}, where the segment is replaced with an alternative observed in the same time window on different days; (ii) \textit{shuffle}, where the same locations are reordered; and (iii) \textit{stationary}, where a single randomly chosen location is repeated across the entire segment duration. Although synthetic, these segments are derived from real behavior to preserve local temporal and spatial plausibility.

%

\subsubsection{Approximate reference construction}
For several abstractions, an ideal construction is not possible. To address this, we employ an approximate approach that leverages the diversity of real trajectories. Our method begins by applying the abstraction $A$ to each trajectory, producing a feature representation that captures the intended behavioral signal. We then cluster these abstractions using an appropriate distance metric (such as the Euclidean distance), grouping users who exhibit similar mobility behaviors. From each cluster, we then select a representative trajectory $T_u$ to serve as the training instance, while all other trajectories in the same cluster form its reference set.

\begin{definition}[Approximate Reference Set]
Let $\{A(T_i)\}_{i=1}^N$ be the abstractions computed over all trajectories in a population. Let $\text{cluster}(\cdot)$ assign trajectories to clusters in the abstraction space. Then, for each medoid trajectory $T_u$ selected from a cluster $\mathcal{C}$, its approximate reference set is defined as:
\[
\tilde{\mathcal{R}} (T_u) = \left\{ T'_v \in \mathcal{T}_{\text{held-out}} \mid \text{cluster}(A(T'_v)) = \text{cluster}(A(T_u)) \right\}.
\]
\end{definition}

This clustering based approach naturally adapts to the population distribution. In our framework, we use it to construct approximate reference sets for the risks of \emph{location memorization} (\cref{subsubsec:location}) and \emph{anchor pair memorization} (\cref{subsubsec:anchor}).

When constructing these sets, we observe that common mobility patterns lead to larger clusters and thus richer reference sets. In contrast, users with more unique patterns may fall into smaller clusters, resulting in limited references and potentially biased memorization scores. To mitigate this, we introduce a cluster enrichment strategy that leverages the fact that trajectories from the held out set can appear in multiple reference sets. Our algorithm (see Appendix \ref{alg:enrichment}) augments small clusters by incorporating nearby neighbors in the abstraction space, subject to a minimum similarity threshold.

\subsection{Memorization Quantification}
\label{subsec:quantify}

Equipped with appropriate reference sets, we can now turn to the measurement of memorization. For this, we follow the design of the exposure metric but reformulate it over the reference set and derive more fine grained statistics from it.
Given a trajectory $T_u$ from the training dataset $\mathcal{T}_{\text{train}}$ of a model and its corresponding  reference set $\mathcal{R}(T_u)$, we define three complementary memorization metrics based on model likelihood $\mathcal{L}(T)$, computed as the negative log-perplexity assigned by the model to $T_u$.

\paragraph{(1) Exposure (\citet{carlini2019secretsharer}).}  
As introduced earlier, exposure measures how highly $T_u$ ranks within its reference set. We can thus formulate it as follows for mobility memorization:
\[
\text{exposure}(T_u)
= \log_2 |\mathcal{R}(T_u)| - \log_2 \operatorname{rank}(\mathcal{L}(T_u)).
\]
In contrast to the original formulation for text, however, the exposure now depends directly on the reference set size $|\mathcal{R}(T_u)|$, making its scale non-uniform across users. As a result, quantitatively comparing memorization for different abstractions and trajectories becomes impossible.



\paragraph{(2) Exposure Preference.}  
This metric captures where $T_u$ stands within the likelihood distribution of its reference set by normalizing its rank into a value in $[0,1]$:
\[
\text{preference}(T_u)
= \frac{\operatorname{rank}(\mathcal{L}(T_u))}{|\mathcal{R}(T_u)|}.
\]
Values close to 1 indicate that the model does not favor the training trajectory over its references (no memorization), whereas values near 0 signal that the model assigns unusually high preference to $T_u$, consistent with memorization.
Unlike exposure, the preference is unaffected by the absolute size of the reference set and therefore supports population-level comparison.

\paragraph{(3) Exposure Magnitude.}  
To capture the \emph{magnitude} of memorization, we compute how much higher the model’s likelihood for $T_u$ is compared to the average alternative:
\[
\text{magnitude}(T_u)
= \mathcal{L}(T_u) - 
\mathbb{E}_{T'_v \sim \mathcal{R}(T_u)}[\mathcal{L}(T'_v)].
\]
Since $\mathcal{L}$ denotes negative log-perplexity, positive values indicate that the model assigns an advantage to the exact training trajectory, consistent with memorization. 
Values near zero reflect appropriate generalization, whereas negative values reflect an absence of memorization and a preference for more typical behavior.


Together, these metrics 
offer complementary perspectives on memorization: ordinal exposure, normalized position, and magnitude, providing a multifaceted evaluation framework.

\section{Experimental Setup}
\label{sec:experimental_setup}
We continue to describe the empirical setup used to evaluate memorization risks in mobility prediction models. This includes the datasets, preprocessing steps, and model configurations employed throughout our experiments. Our aim is to assess memorization across different urban environments and model architectures using real-world human mobility traces.

\subsection{Mobility Datasets}


For our experiments, we consider three public mobility datasets, which are summarized in Table~\ref{tab:dataset_summary}. These datasets span diverse urban contexts and differ in sampling modalities, spatial granularity, and population scale, providing a comprehensive setup for our investigation of mobility memorization.

\paragraph{Shanghai Telecom~\cite{shanghai_kaggle}.}
This dataset records anonymized mobile phone activity from 9,481 users in Shanghai, China, collected via device associations with 3,233 base station. Each entry corresponds to a timestamped connection event (i.e., mobile data), enabling the reconstruction of user mobility over periods ranging from two weeks to six months. The data reflects natural usage patterns in a cellular network and supports long-range trajectory analysis.

\paragraph{YJMob100K~\cite{yjmob100k}.}
This dataset captures fine-grained GPS mobility traces for 100,000 users in a Japanese urban region over a continuous 75-day period. Location points are sampled every 30 minutes and discretized over a uniform 500-meter grid. Each grid cell is enriched with {POI category annotations}, providing both spatial and semantic context. The data was collected through a smartphone application with opt-in user consent.

\paragraph{Shenzhen Urban~\cite{ShenzhenCDRs, ShenzhenUrbanData}.}
Collected from cellular tower logs in Shenzhen, China, this dataset contains over 38 million events for approximately 414,000 users. Each record corresponds to a timestamped network event (call/text message) between a user and one of 1,090 cell towers. Although explicit dates are anonymized, we assume each user is active for a distinct number of days and that the dataset is chronologically ordered per user, following the approach of~\citet{yonga:2025}. This results in a highly skewed distribution of active days, with a median of approximately 9 days, while a minority of users remain active for up to 949 days.


\begin{table}[b]
\centering
\caption{Summary of the mobility datasets used in our experiments, including statistics before and after filtering.}
\label{tab:dataset_summary}
\resizebox{\linewidth}{!}{%
\begin{tabular}{lrrr}
\toprule
Feature & \textbf{Shanghai Telecom} & \textbf{YJMob100K} & \textbf{Shenzhen Urban} \\
\midrule
City & Shanghai, China & Japan & Shenzhen, China \\
\#Records (raw) & 7.2M & 111.5M & 38.2M \\
\#Users (raw) & 9,481 & 100,000 & 414,271 \\
\#Locations (raw) & 3,233 & 34,032 & 1,090 \\
Temporal Span &  up to 180 days & 75 days & up to 949 days \\
\midrule
\#Users (filtered) & 6,132 & 99,610 & 78,591 \\
\#Locations (filtered) & 2,939 & 30,785 & 965 \\
\#Trajectories & 22,993 & 303,779 & 398,440 \\
\bottomrule
\end{tabular}}
\end{table}

\subsection{Preprocessing}
\label{subsec:preprocessing}

To standardize and structure the raw datasets for predictive modeling, we design a uniform preprocessing pipeline that transforms the heterogeneous dataset records into a common sequence format. Summary statistics for this preprocessing are reported in Table~\ref{tab:dataset_summary}. The process comprises four main stages: 

\paragraph{Harmonization.} All datasets are first converted into a normalized format with four fields: \emph{UserID}, \emph{Latitude}, \emph{Longitude}, and \emph{Timestamp}. This ensures compatibility across datasets with varying formats. For sources that only provide time-of-day information (\textit{Shenzhen Urban} and \textit{YJMob100K}), artificial dates are assigned per user starting from day 0 and incremented chronologically.

\paragraph{Discretization.} 
In general, spatial discretization is a necessary step for mobility analysis, particularly when working with raw GPS trajectories, where exact location matches are unlikely due to measurement noise and floating-point precision. Without discretization, trajectories may reflect artificial movement rather than meaningful user behavior, making it difficult to identify recurring locations or periods of stationarity. Therefore, it is standard to apply noise filtering and spatial discretization techniques, such as grid-based tessellation or hierarchical spatial indexing (e.g., H3~\cite{h3geo}), as discussed in~\cite{Zheng2015}.
In our case, the datasets already provide discrete spatial representations (e.g., cell towers or grid-based locations), and no additional spatial discretization is required. 

We then perform temporal discretization to align records to a fixed sampling frequency. All trajectories are resampled at uniform \emph{30-minute intervals}, consistent with the resolution of \textit{YJMob100K}. For each interval, the most frequent location is retained, while missing intervals are handled in a later preprocessing step.



\paragraph{Anchor inference and completion.} 
Each user’s home and work locations are inferred as the most frequently visited places during night hours (23:00–06:00) and work hours (09:00–12:00 and 14:00–18:00), respectively. These inferred anchors are then used to complete missing values during their corresponding time windows. Intervals outside these periods, such as lunch breaks or evening activities, remain unfilled.

\paragraph{User and day filtering.} We keep only users with sufficient activity, defined as having at least 5 valid days, where a day is considered valid if it contains less than 30\% missing intervals. From these, we select each user’s 14 best days (i.e., those with the lowest proportion of missing data) and segment them into non-overlapping trajectories of 3 consecutive days. This choice balances two goals: (i)~providing sufficiently long sequences for effective training of mobility models, and (ii) increasing the number of trajectory samples for downstream memorization analysis, particularly approximate reference set construction.


\subsection{Mobility Prediction Models}
\label{subsec:models}

To evaluate memorization across different setups, we select one representative model from each major architecture used in mobility modeling. These include both classical probabilistic models and state-of-the-art deep learning architectures.

\begin{itemize}[leftmargin=*]
    \setlength{\itemsep}{3pt}
     \item \textit{Markov model~\cite{norris1998markov}.}  
    We implement a classic second-order Markov chain that predicts the next location from the two most recent locations in a user's trajectory. This provides a strong probabilistic baseline capturing short-term transition dynamics.


     \item \textit{Long Short-Term Memory (LSTM)~\cite{hochreiter1997long}.}  
    We implement a standard LSTM-based prediction model (LSTM-simple) that operates on 24-hour input sequences. Locations and timestamps are embedded, concatenated, and passed through the LSTM, followed by a softmax output layer. A variant, LSTM-long, processes the full 72-hour history to examine the effect of longer context.

    \item \textit{DeepMove~\cite{DeepMove}.}  
    This attention-based model splits the trajectory into a long-term history and a current-day segment, using an encoder–decoder structure to capture periodic cross-day patterns.  
    We evaluate two variants: \emph{DM-locallong}, which attends locally from the current-day decoder to a 48h history encoder, and \emph{DM-avglonguser}, which instead pools the entire history into a global context vector and includes a user embedding.  

    \item \textit{LSTPM~\cite{LSTPM}.} This model captures both long-term and short-term mobility preferences. It aggregates trajectories into 48 temporal bins (hour-of-day × weekday/weekend), computes inter-bin similarities to derive long-term preferences, and combines them with a geo-dilated RNN for short-term modeling. The resulting context is used to generate personalized next-location predictions.
    
    \item \textit{Graph-Flashback~\cite{graphflashback}.} This model represents mobility trajectories as a Point-of-Interest (POI) transition graph. A graph convolutional network (GCN) learns POI embeddings based on structural and temporal context. These embeddings are then integrated with user histories to enhance next-location prediction.
\end{itemize}

We use official implementations whenever available\footnote{
DeepMove: \url{https://github.com/vonfeng/DeepMove};  
LSTPM: \url{https://github.com/NLPWM-WHU/LSTPM};  
Graph-Flashback: \url{https://github.com/kevin-xuan/Graph-Flashback}.  
}. Hyperparameters are left largely unchanged from the original repositories; only dataset-dependent parameters (e.g., number of locations) are adapted. All models are trained using a time-based 80–20 train–validation split with early stopping based on validation loss. A complete summary of hyperparameter settings is provided in appendix (cf. Table~\ref{tab:model-hparams}).

\subsection{Experimental Protocol}
\label{subsec:reference-sets}

For the measurement of memorization with our framework, we construct reference sets tailored to each of the three abstractions introduced in \cref{subsec:abstraction}. In all cases, we fix the training set size to $2{,}000$ trajectories, which provides sufficient behavioral diversity for model training while leaving the remaining trajectories available to form large, diverse reference sets for each training sample.

\paragraph{Location and anchor-pair memorization.}
For the first two memorization risks, we construct approximate reference sets by partitioning user behaviors using $k$-means clustering, where we fix the number of clusters to $2{,}000$. Each cluster is represented by one trajectory used for training, while the remaining trajectories form its reference set. To ensure that memorization metrics are computed over sufficiently large and stable samples, we apply a post-processing enrichment step (see \cref{alg:enrichment}). 
This step is governed by two parameters: the target minimum reference set size $k_{\min}$ and a distance threshold $\tau$ controlling similarity. The parameter $k_{\min}$ determines the minimum number of trajectories required in each reference set to obtain statistically reliable estimates, while $\tau$ limits how dissimilar additional trajectories can be when enriching a cluster. These parameters induce a trade-off: smaller $\tau$ values preserve behavioral coherence but may prevent reaching $k_{\min}$, whereas larger values facilitate reaching $k_{\min}$ at the cost of reduced similarity.

In practice, $\tau$ is inherently dataset- and abstraction-dependent, as it depends on the scale and variability of trajectory features, and is therefore not fixed globally. 
Instead, trajectories are added in order of increasing distance until the target size $k_{\min}$ is reached, resulting in an implicit threshold. 
In contrast, $k_{\min}$ should be chosen relative to the dataset size: it must remain small compared to the total number of trajectories to preserve behavioral coherence within each reference set, while being large enough to ensure stable estimation. 
In our datasets (ranging from $\sim$23K to $\sim$398K trajectories), we set $k_{\min}=100$ as a rule of thumb that provides a good balance between these objectives.
Under this setting, reference set sizes vary significantly across datasets (from 100 to 144,891; see Figure~\ref{fig:refset_sizes} in appendix), and the resulting effective values of $\tau$ also differ across memorization risks (e.g., $\tau \approx 1.67$ for location memorization in Shenzhen versus $\tau \approx 45.31$ for anchor-pair normalization in YJMob100K). Finally, to ensure that clusters capture shared behavioral patterns rather than user identity, we analyze the spatial dispersion between cluster representatives and their reference sets (Figure~\ref{fig:refset_geo_dispersion} in appendix), confirming that reference trajectories remain sufficiently varied while still reflecting consistent mobility behaviors.


\paragraph{Segment-level memorization.}
To evaluate segment-level memorization, we generate controlled variants of real trajectories. 
We again sample $2{,}000$ trajectories for training and construct reference sets by perturbing short temporal segments. 
Each trajectory is divided into non-overlapping 4-hour windows (00:00–20:00), providing multiple localized positions where alternative behavior can be inserted. 
A short window is chosen to preserve realism, as larger perturbation spans tend to produce implausible mobility patterns. 
For each window, we generate synthetic references using three transformations---\textit{shuffle}, \textit{stationary}, and \textit{substitute} (cf. \cref{subsec:reference})---and balance them to obtain 300 references per trajectory.


\paragraph{Model performance.}

The resulting training subsets are then used to learn mobility prediction models with the hyperparameter settings specified in Table~\ref{tab:model-hparams}.
We report the corresponding predictive performance in Table~\ref{tab:accuracy}.

Overall, the models exhibit comparable accuracy trends across datasets, with \textit{DM-locallong} achieving the strongest performance and \textit{Markov} remaining surprisingly competitive despite its simplicity.
One exception is \textit{Graph-Flashback}, whose accuracy is substantially lower. This behavior is consistent with the original architecture, which relies on relatively small embedding and hidden dimensions (10 units), limiting its capacity to capture complex mobility patterns. While its performance on smaller datasets remains within the range reported in prior work~\cite{graphflashback}, it degrades on larger and more diverse datasets such as \textit{ShanghaiTel} and \textit{YJMob100K}, where trajectory variability is higher, following the general trend of decreasing accuracy with increasing location entropy (see Table~\ref{tab:dataset_summary}).
Importantly, this lower predictive performance provides a useful contrast, allowing us to study memorization under a lower-capacity regime where the model captures less global structure and may rely differently on memorization. As such, Graph-Flashback serves as an informative outlier highlighting how memorization behaviors vary across models with different representational capacities.

These results confirm that models capture mobility structure to varying degrees while being far from highly optimized; a setting well suited for analyzing memorization behavior across architectures with different capacities.

\newcommand{\accMarkovSZTOne}{91.3}
\newcommand{\accMarkovSZTFive}{92.3}
\newcommand{\accMarkovSHTOne}{76.2}
\newcommand{\accMarkovSHTFive}{79.7}
\newcommand{\accMarkovYJTOne}{22.5}
\newcommand{\accMarkovYJTFive}{38.1}
\newcommand{\accLSTMsimpleSZTOne}{89.7}
\newcommand{\accLSTMsimpleSZTFive}{92.8}
\newcommand{\accLSTMsimpleSHTOne}{73.2}
\newcommand{\accLSTMsimpleSHTFive}{80.2}
\newcommand{\accLSTMsimpleYJTOne}{18.2}
\newcommand{\accLSTMsimpleYJTFive}{22.9}
\newcommand{\accLSTMlongSZTOne}{90.0}
\newcommand{\accLSTMlongSZTFive}{93.4}
\newcommand{\accLSTMlongSHTOne}{78.0}
\newcommand{\accLSTMlongSHTFive}{84.8}
\newcommand{\accLSTMlongYJTOne}{18.8}
\newcommand{\accLSTMlongYJTFive}{24.5}
\newcommand{\accDeepMovelocallongSZTOne}{92.7}
\newcommand{\accDeepMovelocallongSZTFive}{96.0}
\newcommand{\accDeepMovelocallongSHTOne}{84.6}
\newcommand{\accDeepMovelocallongSHTFive}{87.6}
\newcommand{\accDeepMovelocallongYJTOne}{20.2}
\newcommand{\accDeepMovelocallongYJTFive}{25.3}
\newcommand{\accDeepMoveavglonguserSZTOne}{92.4}
\newcommand{\accDeepMoveavglonguserSZTFive}{95.0}
\newcommand{\accDeepMoveavglonguserSHTOne}{71.4}
\newcommand{\accDeepMoveavglonguserSHTFive}{79.2}
\newcommand{\accDeepMoveavglonguserYJTOne}{18.1}
\newcommand{\accDeepMoveavglonguserYJTFive}{22.9}
\newcommand{\accLSTPMSZTOne}{86.5}
\newcommand{\accLSTPMSZTFive}{92.7}
\newcommand{\accLSTPMSHTOne}{58.6}
\newcommand{\accLSTPMSHTFive}{68.2}
\newcommand{\accLSTPMYJTOne}{15.5}
\newcommand{\accLSTPMYJTFive}{20.4}
\newcommand{\accGraphFlashbackSZTOne}{32.0}
\newcommand{\accGraphFlashbackSZTFive}{57.2}
\newcommand{\accGraphFlashbackSHTOne}{10.4}
\newcommand{\accGraphFlashbackSHTFive}{23.2}
\newcommand{\accGraphFlashbackYJTOne}{00.3}
\newcommand{\accGraphFlashbackYJTFive}{00.7}

\begin{table}[h]
\centering
\small
\caption{Average (top-1 and top-5) models' accuracy.}
\label{tab:accuracy}
\begin{tabular}{lcccccc}
\toprule
Model & \multicolumn{2}{c}{\textbf{ShenzhenUrb.}} & \multicolumn{2}{c}{\textbf{ShanghaiTel.}} & \multicolumn{2}{c}{\textbf{YJMob100K}} \\
\cmidrule(lr){2-3} \cmidrule(lr){4-5} \cmidrule(lr){6-7}
& Top-1 & Top-5 & Top-1 & Top-5 & Top-1 & Top-5 \\
\midrule
Markov (2nd)           & \accMarkovSZTOne           & \accMarkovSZTFive           & \accMarkovSHTOne           & \accMarkovSHTFive           & \accMarkovYJTOne           & \accMarkovYJTFive           \\
LSTM-simple            & \accLSTMsimpleSZTOne       & \accLSTMsimpleSZTFive       & \accLSTMsimpleSHTOne       & \accLSTMsimpleSHTFive       & \accLSTMsimpleYJTOne       & \accLSTMsimpleYJTFive       \\
LSTM-long              & \accLSTMlongSZTOne         & \accLSTMlongSZTFive         & \accLSTMlongSHTOne         & \accLSTMlongSHTFive         & \accLSTMlongYJTOne         & \accLSTMlongYJTFive         \\
DM-locallong     & \accDeepMovelocallongSZTOne   & \accDeepMovelocallongSZTFive   & \accDeepMovelocallongSHTOne   & \accDeepMovelocallongSHTFive   & \accDeepMovelocallongYJTOne   & \accDeepMovelocallongYJTFive   \\
DM-avglonguser   & \accDeepMoveavglonguserSZTOne & \accDeepMoveavglonguserSZTFive & \accDeepMoveavglonguserSHTOne & \accDeepMoveavglonguserSHTFive & \accDeepMoveavglonguserYJTOne & \accDeepMoveavglonguserYJTFive \\
LSTPM                  & \accLSTPMSZTOne            & \accLSTPMSZTFive            & \accLSTPMSHTOne            & \accLSTPMSHTFive            & \accLSTPMYJTOne            & \accLSTPMYJTFive            \\
Graph-Flashback        & \accGraphFlashbackSZTOne   & \accGraphFlashbackSZTFive   & \accGraphFlashbackSHTOne   & \accGraphFlashbackSHTFive   & \accGraphFlashbackYJTOne   & \accGraphFlashbackYJTFive   \\
\bottomrule
\end{tabular}
\end{table}

\section{Empirical Findings}
\label{sec:empirical_findings}


We structure our discussing of findings along the research questions formulated in the introduction. In particular, we explore whether memorization occurs (RQ1: \cref{subsec:rq1}), what best explains it (RQ2:~\cref{subsec:rq2}), how it varies across models (RQ3:~\cref{subsec:rq3}), and how it relates to extractability~(RQ4:~\cref{subsec:rq4}). Each question is addressed under controlled conditions by fixing key factors such as model, dataset, or memorization risk and varying a single dimension to isolate its effect.

\newcommand{\SZposAdv}{99.4\%}
\newcommand{\SZadvMean}{0.74}
\newcommand{\SZexpMed}{4.19}
\newcommand{\SZpreference}{67\%}

\newcommand{\SHposAdv}{87.1\%}
\newcommand{\SHadvMed}{1.16}
\newcommand{\SHexpMed}{1.79}
\newcommand{\SHpreference}{21.8\%}

\newcommand{\YJposAdv}{99.95\%}
\newcommand{\YJadvMed}{5.95}
\newcommand{\YJexpMed}{5.64}
\newcommand{\YJpreference}{88.9\%}

\subsection{Detecting Individual Memorization}
\label{subsec:rq1}

We begin by assessing whether mobility memorization arises when training on our three datasets. For this experiment, we focus on the model \emph{LSTM-simple}, as it provides a simple and interpretable neural baseline for analyzing memorization behavior.
Experiments involving all models are discussed later in \cref{subsec:rq3}. We train LSTM-simple on each dataset and compute the proposed memorization metrics under the \textit{location memorization} abstraction. The training dynamics are shown in \cref{fig:app_training_dynamics} in the appendix, where we observe a continuous decrease in loss, indicating that overfitting has not occurred.

\begin{figure}
\centering
\begin{subfigure}{0.32\linewidth}
    \centering
    \includegraphics[width=\linewidth]{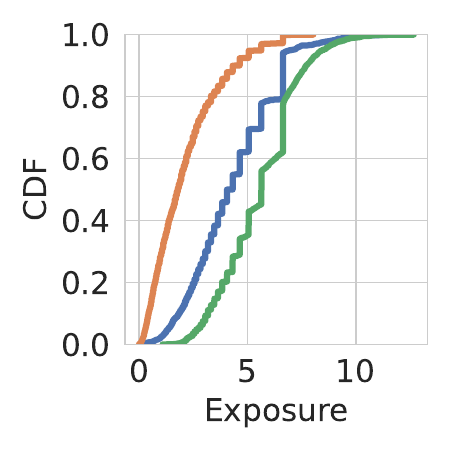}
    \caption{Exposure CDF}
    \label{fig:rq1_exposure_cdf}
\end{subfigure}
\hfill
\begin{subfigure}{0.32\linewidth}
    \centering
    \includegraphics[width=\linewidth]{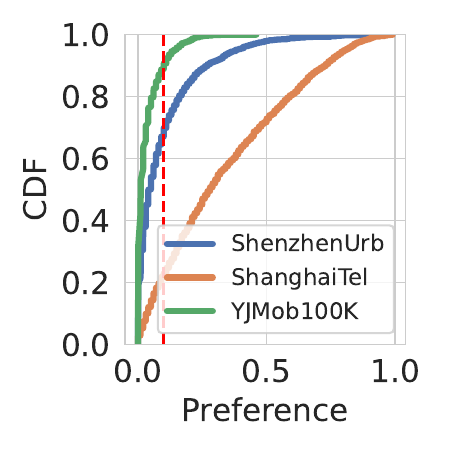}
    \caption{Preference CDF}
    \label{fig:rq1_percentile_cdf}
\end{subfigure}
\hfill
\begin{subfigure}{0.32\linewidth}
    \centering
    \includegraphics[width=\linewidth]{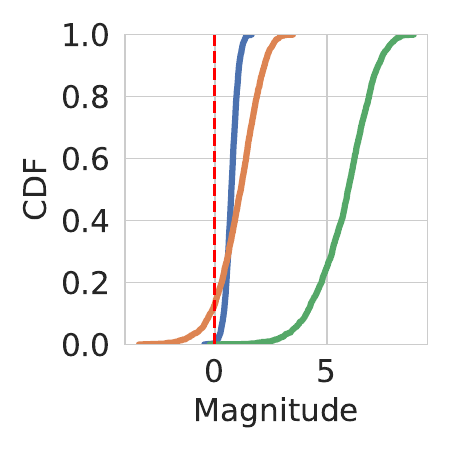}
    \caption{Magnitude CDF}
    \label{fig:rq1_gap_cdf}
\end{subfigure}
\caption{Cumulative distribution of mobility memorization metrics for the \textit{LSTM-simple} model across datasets.}
\label{fig:rq1_memorization_metrics}
\end{figure}

Figure~\ref{fig:rq1_memorization_metrics} shows the distributions of memorization metrics for all training trajectories. Across the three datasets, we observe clear and systematic memorization signals.


\paragraph{Memorization on ShenzhenUrb}  
Memorization is strongest for this dataset with low mobility entropy.  
Over \SZposAdv{} of trajectories have positive exposure magnitude, with a mean of \SZadvMean{}.  Exposure values are high (median \SZexpMed{}), and \SZpreference{} of trajectories stand within the bottom 10\% of their reference likelihood distribution, showing strong preferential treatment for true training sequences.

\paragraph{Memorization on ShanghaiTel}  
For this dataset, patterns are qualitatively similar but attenuated.  
Exposure magnitude remains positive for \SHposAdv{} of trajectories (median \SHadvMed{}).  
Exposure scores decrease substantially (median \SHexpMed{}), and only \SHpreference{} of trajectories appear in the bottom 10\% preference region, indicating weaker but still non-negligible memorization.

\paragraph{Memorization on YJMob100K}  
Despite being the most heterogeneous dataset, memorization manifests strongly.  
Exposure magnitude is positive for \YJposAdv{} of trajectories, with a large median of \YJadvMed{}.  
Exposure remains high (median \YJexpMed{}), and a striking \YJpreference{} of trajectories fall within the most confident 10\% of preference scores, revealing a strong model tendency to favor true sequences.

\medskip 
\finding{ Our findings demonstrate that mobility memorization arises across datasets of varying scale and entropy. Even without overfitting, prediction models tend to retain user-specific locations rather than generalize from behavior. 
}

\begin{figure*}
\centering
\begin{subfigure}{0.32\linewidth}
    \centering
    \includegraphics[width=\linewidth]{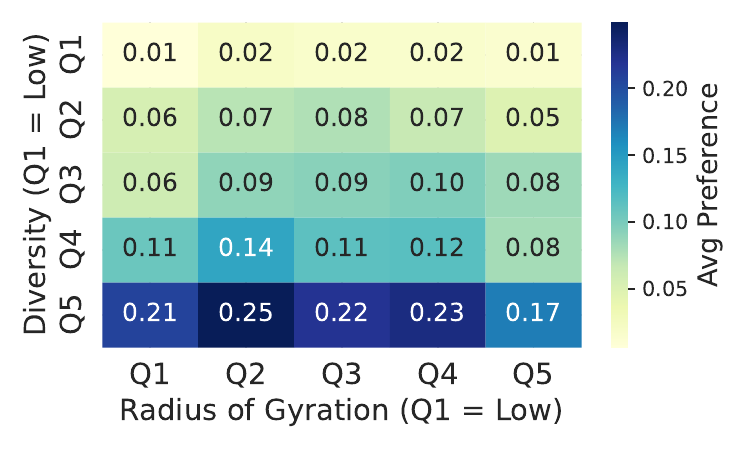}
    \caption{Preference (div vs. rg)}
    \label{fig:rq2_heatmap_rg_diversity}
\end{subfigure}
\hfill
\begin{subfigure}{0.25\textwidth}
    \centering
    \includegraphics[width=\linewidth]{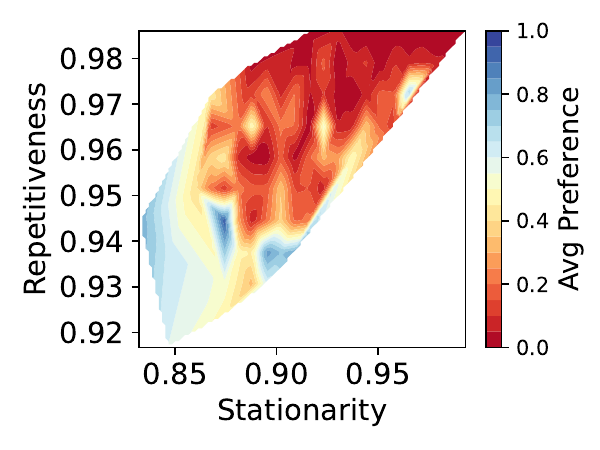}
    \caption{Preference (sta vs. rep)}
    \label{fig:rq2_contour_stationarity_repetitiveness}
\end{subfigure}
\hfill
\begin{subfigure}{0.22\textwidth}
    \centering
    \includegraphics[width=\linewidth]{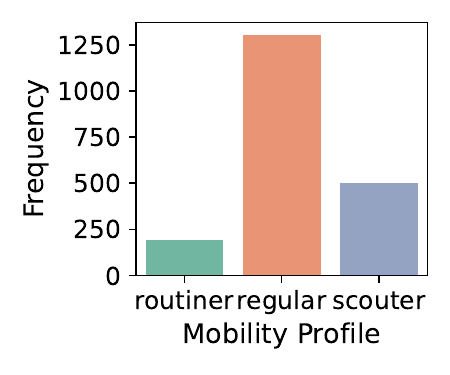}
    \caption{Mobility profiles.}
    \label{fig:rq2_profile_histogram}
\end{subfigure}
\hfill
\begin{subfigure}{0.19\textwidth}
    \centering
    \includegraphics[width=\linewidth]{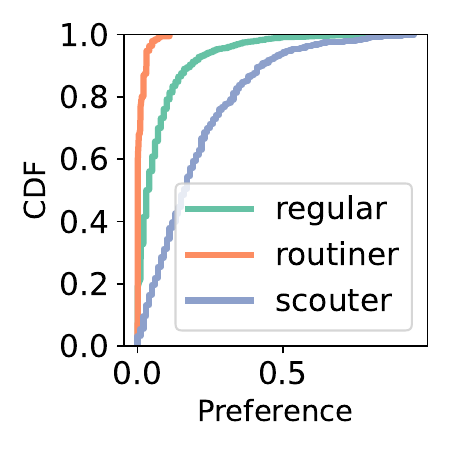}
    \caption{Preference (profiles).}
    \label{fig:rq2_ridge_percentile_by_profile}
\end{subfigure}
\caption{
Mobility characteristics and memorization patterns for \textit{LSTM-simple} on the \textit{Shenzhen Urban} dataset.
(a) Heatmap of memorization preference as a function of diversity (div) and radius of gyration (rg); lighter colors indicate lower preference and thus higher memorization.
(b) Contour plot of memorization preference with respect to stationarity (sta) and repetitiveness (rep); cooler (blue) regions correspond to lower memorization (higher preference). 
(c) Distribution of user mobility profiles (frequency of trajectories per profile).
(d) Cumulative distribution of memorization preference across profiles.
}
\label{fig:rq2_memorization_vs_mobility}
\end{figure*}

\begin{figure*}
\centering
\begin{subfigure}{0.23\textwidth}
    \centering
    
    \includegraphics[width=\linewidth]{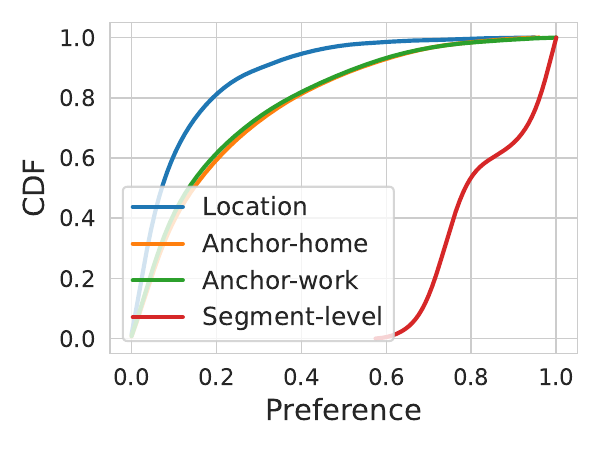}
    \caption{Preference across memorization types.}
    \label{fig:rq3_cdf_percentile_by_type}
\end{subfigure}
\hfill
\begin{subfigure}{0.23\textwidth}
    \centering
    \includegraphics[width=\linewidth]{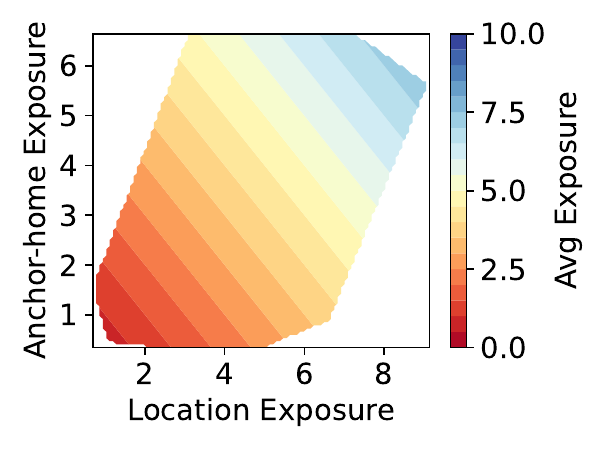}
    \caption{Exposure comparison: Location vs. Anchor-home.}
    \label{fig:rq3_contour_location_vs_home}
\end{subfigure}
\hfill
\begin{subfigure}{0.23\textwidth}
    \centering
    \includegraphics[width=\linewidth]{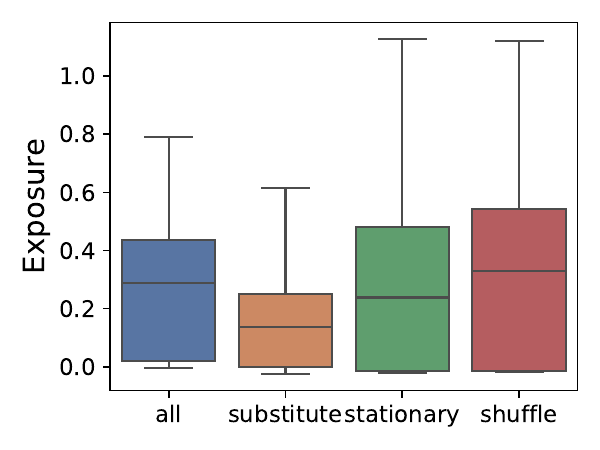}
    \caption{Exposure variation across transformations (Segment-level).}
    \label{fig:rq3_boxplot_exposure_by_mode}
\end{subfigure}
\hfill
\begin{subfigure}{0.23\textwidth}
    \centering
    \includegraphics[width=\linewidth]{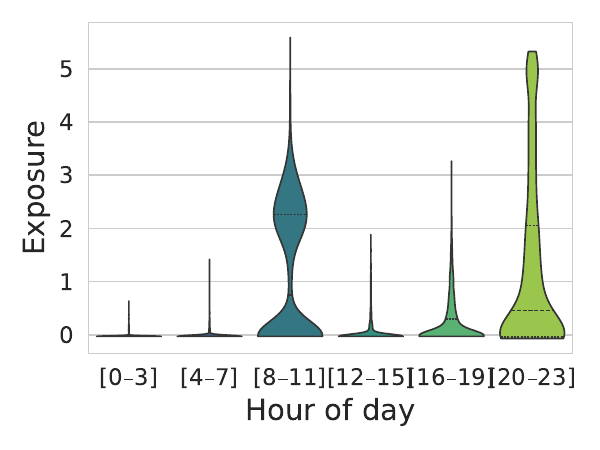}
    \caption{Exposure by time-of-day segments (Segment-level).}
    \label{fig:rq3_violin_exposure_by_hour}
\end{subfigure}
\caption{
Memorization behavior across risks, transformations, and sub-trajectories (\textit{LSTM-simple}, \textit{Shenzhen Urban}).
(a) Cumulative distribution of preference across memorization types.
(b) Contour plot comparing exposure between location-level and anchor-home memorization; cooler (blue) regions indicate higher exposure (and thus higher memorization).
(c) Boxplot of exposure across segment-level transformations; boxes represent the interquartile range (IQR), the central line denotes the median, and whiskers extend to 1.5$\times$IQR.
(d) Violin plot of exposure across time-of-day segments; the width of each violin reflects the density of trajectories, with wider sections indicating more frequent exposure values, while central markers indicate median values.
Higher exposure corresponds to higher memorization.
}
\label{fig:rq3_memorization_patterns}
\end{figure*}

\subsection{Who is Memorized and Why}
\label{subsec:rq2}

To understand which users are more likely to be memorized by predictive models, we analyze how their individual behavior correlates with the memorization metrics. To obtain a clearer grasp of this behavior, we focus on four well-established features known to capture key properties of human movement~\cite{teixeira:hal-03360537, mucceli:2016, Gonzalez:2008, Amichi:2020}:
\begin{itemize}
    \setlength{\itemsep}{3pt}
    \item[rep] \textit{repetitiveness}, defined as the fraction of returns to previously visited locations within a trajectory.
    \item[sta] \textit{stationarity}, the proportion of time spent at the same location across consecutive intervals, indicating sedentary behavior.
    \item[div] \textit{diversity}, measured as the number of distinct sub trajectories, reflecting spatial variability.
    \item[rg]  \textit{radius of gyration}, which quantifies the spread of movement, with higher values indicating broader coverage.
\end{itemize}

Based on these features, the users are grouped into three behavioral mobility profiles, i.e., \textit{routiner}, \textit{regular}, and \textit{scouter}, which represent decreasing levels of spatial and temporal regularity~\cite{Amichi:2020}. Further details on this analysis and the used features is presented in \cref{appendix:mobility_metrics} of the appendix.
%
When investigating the Shenzhen Urban dataset with LSTM-simple, we observe consistent and interpretable patterns, as shown in \cref{fig:rq2_memorization_vs_mobility}.
Corresponding analyses for other datasets are discussed in the Appendix \ref{appendix:rq2_details} (\cref{fig:rq2_memorization_vs_mobility_yjmob}), with similar structural trends but differences in magnitude and distribution.

\paragraph{Radius of gyration.} First, we find that users with low diversity (Q1 < 0.6) and small radius of gyration (Q1 < 10.73~km) are most memorized, with exposure preferences concentrated near zero (see \cref{fig:rq2_heatmap_rg_diversity}).
Interestingly, we find that the radius of gyration exhibits a U-shaped effect. Users with either very low (Q1) or very high (Q4 > 20.82 km) gyration values tend to be memorized more, while those with mid-range values are less retained. This latter trend, however, is not consistent across all datasets (see appendix \cref{fig:rq2_memorization_vs_mobility_yjmob}).

\paragraph{Stationarity and repetitiveness} Second, we observe that memorization generally increases with both stationarity and repetitiveness. Users who tend to stay in the same place and frequently return to previously visited locations during movememnts are more likely to be memorized compared to users with less stationary and less repetitive activities (see \cref{fig:rq2_contour_stationarity_repetitiveness}).

\paragraph{Mobility profiles} Finally, similar effects are also reflected in  the mobility profiles:
\cref{fig:rq2_profile_histogram} and \cref{fig:rq2_ridge_percentile_by_profile} show that \textit{routiners} (9.5\%), characterized by high stationarity and low diversity, tend to experience stronger memorization,
with memorization exposure preferences tightly concentrated below 0.1. They are followed by \textit{regular} users (65.3\%), who show moderate retention. In contrast, \textit{scouters} (25.2\%), who demonstrate greater spatial and temporal variability, are memorized far less, with exposure preference distributions skewed higher, indicating lower exposure.

\medskip
\finding{
Our findings reveal that memorization in predictive mobility models is far from uniform. 
It tends to disproportionately affect users with regular, repetitive, and localized mobility patterns, which are easier to learn and thus more likely to be memorized.
}


\subsection{Memorization Across Risks and Models}
\label{subsec:rq3}

To further explore how memorization behaves under different settings, we investigate how our metrics vary across abstractions and model architectures. Using the Shenzhen Urban dataset, we again show our analysis on the LSTM-simple model to isolate the effects of varying abstractions and generalize later across model variants. 
Equivalent analyses for other model architectures are discussed in Appendix \ref{appendix:rq3_details} (\cref{fig:rq3_memorization_patterns_appendix}), with similar trends but variations in magnitude and consistency across models.

\paragraph{Memorization risk varies by abstraction.}
\cref{fig:rq3_cdf_percentile_by_type} shows that the same model exhibits different levels of memorization depending on the abstracted mobility pattern. We observe {stronger} memorization for location sequences, indicated by a highly skewed distribution toward low preference values (with 80\% lower than 0.2); {more} moderate memorization for anchor pairs, with exposure preferences more evenly spread; and 
{much lower memorization for segment-level patterns}, where the preference values cluster around 0.6–1.0. 

This contrast between abstractions holds despite varying degrees of abstraction, i.e., the extent to which each abstraction modifies the input sequence: full-trajectory changes for location-based memorization, partial windows (e.g., morning or evening hours) for anchor-based memorization, and short sub-sequences (e.g., 4-hour slices) for segment-level analysis. 
{Similar patterns are generally observed across other model architectures (\cref{fig:rq3_memorization_patterns_appendix}), although the magnitude of the differences varies and segment-level memorization is not uniformly negligible.}


\paragraph{Anchor-pair and location memorization are correlated.}
\cref{fig:rq3_contour_location_vs_home} compares user-level memorization exposure between anchor-pair (home) and location-based settings, focusing on the 109 users present in both training subsets. Despite the limited overlap, we observe 
{a positive relationship}: users who are memorized in the location-based setting tend to also be memorized under anchor-based conditions. 
{Comparable cross-abstraction relationships are also observed for other architectures
}, as reported in Figure~\ref{fig:rq3_memorization_patterns_appendix} in the appendix.
A similar trend holds between home-based and work-based anchor pair memorization, see Figure~\ref{fig:rq3_contour_home_vs_work_all_models} in the appendix.


\paragraph{Transformation matter in segment-level memorization}
We observe that the nature of transformations influences memorization. Figure~\ref{fig:rq3_boxplot_exposure_by_mode} compares different transformations at the segment level and shows that exposure generally remains low, yet the form of transformation matters: \textit{shuffle} introduces the most deviation, yielding the highest exposure values. This suggests that the model struggles slightly more when sequences are reordered unexpectedly. In contrast, \textit{substitute} and \textit{stationary} modes yield lower exposure, reflecting the model's ability to tolerate more semantically plausible changes, such as swapping a routine with another similar one or remaining at a single location. 
{This ranking is observed in several models, although the differences between transformations can be smaller or partially overlapping depending on the architecture.} 


In Figure~\ref{fig:rq3_violin_exposure_by_hour}, we further explore how memorization varies by the time of day at which the transformation occurs. The distribution reveals almost no exposure during nighttime hours (0–7h), where user activity is minimal and patterns are uniform. By contrast, exposure {exposure tends to be higher} 
during two windows: early morning (8–11h) and late evening (20–23h). These intervals often capture transitions such as commuting or post-work activities, where user routines may be less consistent. Interestingly, exposure values during these periods can exceed 5 in some cases, despite segment-level memorization generally being low, indicating that for some users, the model retains highly specific behaviors even in varying, non-routine parts of the day. 
{Similar temporal tendencies can be observed in some architectures (e.g., DM-avglonguser), although the effect is less pronounced or absent in others (Figure~\ref{fig:rq3_memorization_patterns_appendix}).}

\medskip
\finding{
{Memorization tends to concentrate on long-term and routine-based behaviors, while short segments are generally less memorized.  Users memorized under one abstraction often remain vulnerable under others, although the strength of this effect varies across models. 
Segment-level memorization can still occur during transition periods, where user behavior is more distinctive.}}

\begin{figure}
    \centering
    \includegraphics[width=\linewidth]{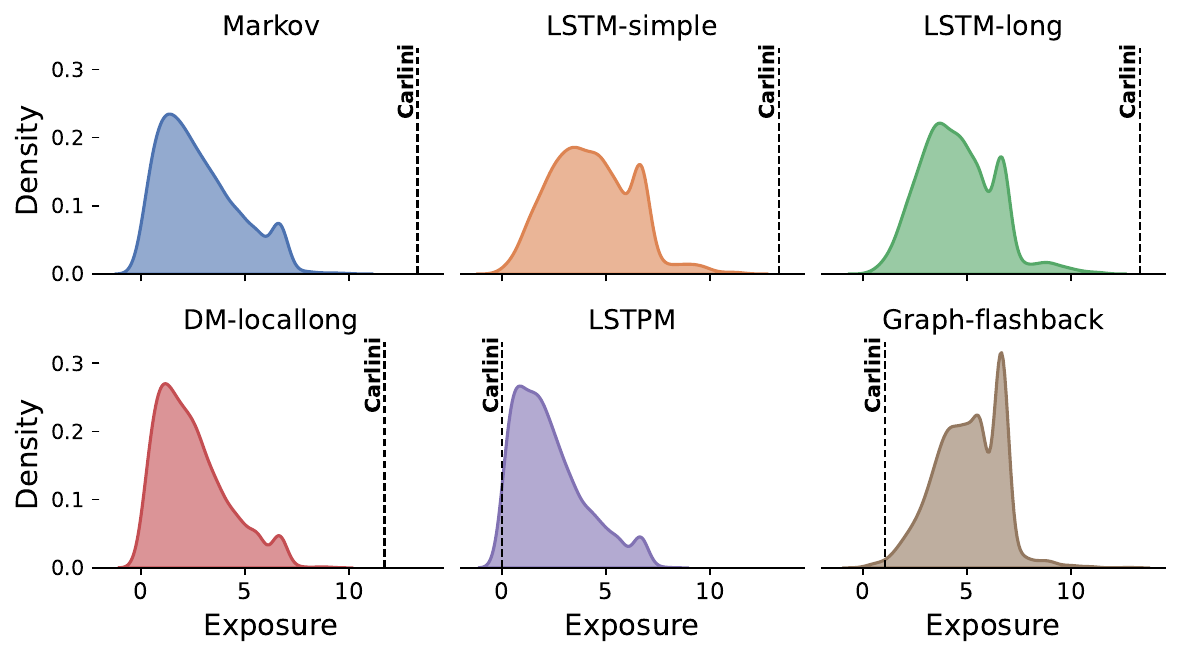}
    \caption{
Exposure distributions across models on the \textit{Shenzhen Urban} dataset, with exposure markers from \citet{carlini2019secretsharer}.
{Each curve represents a kernel density estimate of exposure values across trajectories. 
The shape of the density reflects how exposure is distributed within a model: peaks indicate common exposure levels, while spread and tails capture variability and the presence of high memorization. 
}}
    \label{fig:enter-label}
\end{figure}


\newcommand{\markovExpMean}{2.82}
\newcommand{\markovExpStd}{1.87}
\newcommand{\markovPrefMean}{0.25}
\newcommand{\markovPrefStd}{0.24}
\newcommand{\markovMagMean}{2.97}
\newcommand{\markovMagStd}{2.10}

\newcommand{\lstmSimpleExpMean}{4.31}
\newcommand{\lstmSimpleExpStd}{1.95}
\newcommand{\lstmSimplePrefMean}{0.10}
\newcommand{\lstmSimplePrefStd}{0.13}
\newcommand{\lstmSimpleMagMean}{0.74}
\newcommand{\lstmSimpleMagStd}{0.28}

\newcommand{\lstmLongExpMean}{4.57}
\newcommand{\lstmLongExpStd}{1.77}
\newcommand{\lstmLongPrefMean}{0.07}
\newcommand{\lstmLongPrefStd}{0.09}
\newcommand{\lstmLongMagMean}{0.77}
\newcommand{\lstmLongMagStd}{0.24}

\newcommand{\dmLocalExpMean}{2.49}
\newcommand{\dmLocalExpStd}{1.68}
\newcommand{\dmLocalPrefMean}{0.28}
\newcommand{\dmLocalPrefStd}{0.23}
\newcommand{\dmLocalMagMean}{0.47}
\newcommand{\dmLocalMagStd}{0.31}

\newcommand{\dmAvgExpMean}{3.02}
\newcommand{\dmAvgExpStd}{1.87}
\newcommand{\dmAvgPrefMean}{0.22}
\newcommand{\dmAvgPrefStd}{0.22}
\newcommand{\dmAvgMagMean}{0.49}
\newcommand{\dmAvgMagStd}{0.26}

\newcommand{\lstpmExpMean}{2.31}
\newcommand{\lstpmExpStd}{1.69}
\newcommand{\lstpmPrefMean}{0.32}
\newcommand{\lstpmPrefStd}{0.26}
\newcommand{\lstpmMagMean}{0.63}
\newcommand{\lstpmMagStd}{0.68}

\newcommand{\gfExpMean}{5.07}
\newcommand{\gfExpStd}{1.57}
\newcommand{\gfPrefMean}{0.05}
\newcommand{\gfPrefStd}{0.08}
\newcommand{\gfMagMean}{469.15}
\newcommand{\gfMagStd}{1230.36}

\begin{table}[b]
\centering
\small
\caption{Memorization metrics (mean $\pm$ std) for all models on {Shenzhen Urban} under the \textit{location} abstraction.}
\label{tab:memorization-metrics-shenzhen}
\begin{tabular}{lccc}
\toprule
Model & Exposure & Preference & Magnitude \\
\midrule
Markov 
    & \markovExpMean{} $\pm$ \markovExpStd{}
    & \markovPrefMean{} $\pm$ \markovPrefStd{}
    & \markovMagMean{} $\pm$ \markovMagStd{} \\
LSTM-simple 
    & \lstmSimpleExpMean{} $\pm$ \lstmSimpleExpStd{}
    & \lstmSimplePrefMean{} $\pm$ \lstmSimplePrefStd{}
    & \lstmSimpleMagMean{} $\pm$ \lstmSimpleMagStd{} \\
LSTM-long 
    & \lstmLongExpMean{} $\pm$ \lstmLongExpStd{}
    & \lstmLongPrefMean{} $\pm$ \lstmLongPrefStd{}
    & \lstmLongMagMean{} $\pm$ \lstmLongMagStd{} \\
DM-locallong 
    & \dmLocalExpMean{} $\pm$ \dmLocalExpStd{}
    & \dmLocalPrefMean{} $\pm$ \dmLocalPrefStd{}
    & \dmLocalMagMean{} $\pm$ \dmLocalMagStd{} \\
DM-avglonguser 
    & \dmAvgExpMean{} $\pm$ \dmAvgExpStd{}
    & \dmAvgPrefMean{} $\pm$ \dmAvgPrefStd{}
    & \dmAvgMagMean{} $\pm$ \dmAvgMagStd{} \\
LSTPM 
    & \lstpmExpMean{} $\pm$ \lstpmExpStd{}
    & \lstpmPrefMean{} $\pm$ \lstpmPrefStd{}
    & \lstpmMagMean{} $\pm$ \lstpmMagStd{} \\
Graph-Flashback 
    & \gfExpMean{} $\pm$ \gfExpStd{}
    & \gfPrefMean{} $\pm$ \gfPrefStd{}
    & \gfMagMean{} $\pm$ \gfMagStd{} \\
\bottomrule
\end{tabular}
\end{table}

\paragraph{Model architectures yield distinct memorization patterns.}
We now examine how memorization varies across model architectures under the \textit{location} abstraction. Results for \emph{ShenzhenUrban} are shown in Figure~\ref{fig:enter-label}, with analogous patterns observed for \emph{ShanghaiKaggle} and \emph{YJMob100K} (see Figure~\ref{fig:rq4_all_ridgeplots_appendix} {in appendix}). Additionally, Table \ref{tab:memorization-metrics-shenzhen} additionally shows all metrics for all models.

{Exposure distributions differ across models in both spread and tail behavior.}
\emph{LSTPM} and \emph{DM-locallong} exhibit the lowest memorization, with compact distributions concentrated near small exposure values, whereas \emph{Graph-flashback} displays a much heavier tail and substantial mass beyond exposure~5. This is noteworthy given its very small model capacity (10-dimensional embeddings and hidden states), suggesting that reduced representational power does not prevent strong memorization effects.
Even minor architectural changes can shift memorization: \emph{LSTM-simple} and \emph{LSTM-long} share similar shapes, yet extending the temporal input from 24h to 72h noticeably increases the upper tail of the distribution, indicating that longer input horizons can raise memorization pressure.
The \emph{Markov} baseline falls mid-range between neural models, demonstrating that memorization can arise even in simple models. 


\paragraph{Adapted vs. original exposure.}
In addition, we reproduce the original  exposure protocol by \citet{carlini2019secretsharer} as a benchmark: one synthetic “canary” trajectory is inserted into the training set, while 10,000 others—generated by uniformly sampling locations with equal probability—serve as references during testing. This yields a model-level exposure score based on how strongly the model ranks the true canary over the references. However, as shown in Figure~\ref{fig:enter-label}, these Carlini scores (dashed lines) do not consistently align with user-level memorization. In some cases (e.g., \emph{LSTPM}), the Carlini score sits near the lower end of the user-level distribution, whereas in others (e.g., \emph{LSTM-simple}, \emph{DM-locallong}), it exceeds real trajectory scores substantially. 

\medskip
\finding{Memorization depends jointly on architectural design, temporal context, and model capacity. That is, small hyperparameter changes can shift memorization, and small models memorize as well. Moreover, while \citet{carlini2019secretsharer}’s synthetic canary metric captures extreme cases, it does not align with how real user trajectories are memorized across architectures.}

\begin{figure}[t]
\centering
\begin{subfigure}{0.32\linewidth}
    \includegraphics[width=\linewidth]{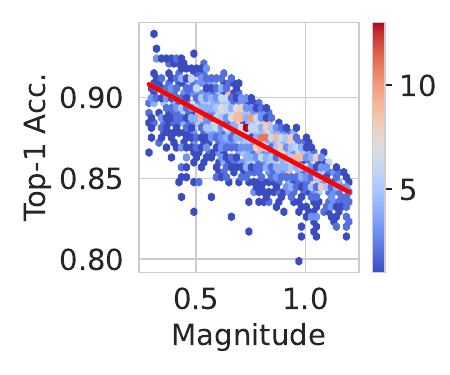}
    \caption{{Shenzhen Urban}}
    \label{fig:rq4_hexbin_gap_top1_shenzhen}
\end{subfigure}
\hfill
\begin{subfigure}{0.32\linewidth}
    \includegraphics[width=\linewidth]{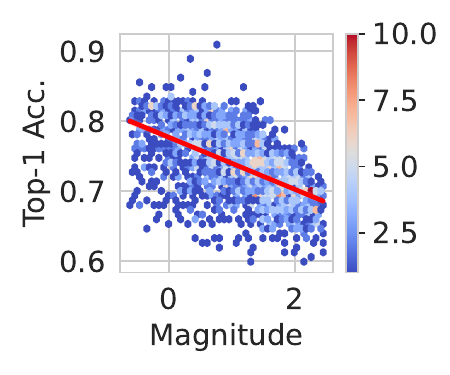}
    \caption{{Shanghai Telecom}}
    \label{fig:rq4_hexbin_gap_top1_shanghai}
\end{subfigure}
\hfill
\begin{subfigure}{0.32\linewidth}
    \includegraphics[width=\linewidth]{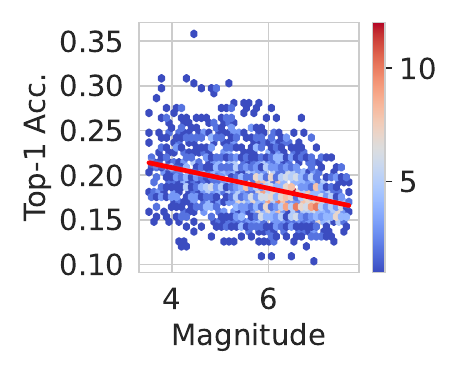}
    \caption{{YJMob100K}}
    \label{fig:rq4_hexbin_gap_top1_yjmob}
\end{subfigure}
\caption{
Utility (Top-1 Accuracy) vs.\ Memorization (Magnitude) across datasets using \textit{LSTM-simple}. 
{Each hexagonal bin aggregates trajectories with similar values, and the color scale indicates the number of trajectories within each bin, with warmer colors corresponding to higher concentrations. }
}
\label{fig:rq4_memorization_vs_utility_hexbin}
\end{figure}
\begin{figure}[t]
\centering
\begin{subfigure}{0.48\linewidth}
    \includegraphics[width=\linewidth]{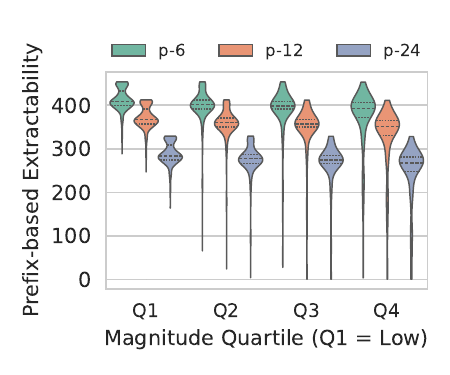}
    \caption{Prefix-based extractability.}
    \label{fig:rq4_violin_gap_prefix}
\end{subfigure}
\hfill
\begin{subfigure}{0.48\linewidth}
    \includegraphics[width=\linewidth]{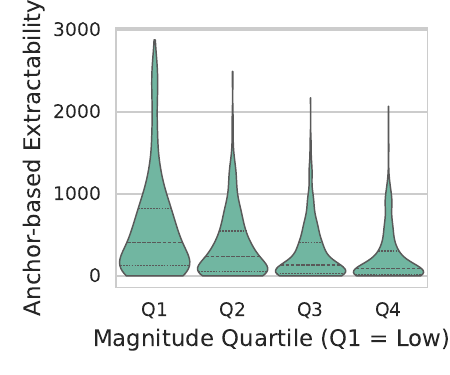}
    \caption{Anchor-based extractability.}
    \label{fig:rq4_violin_gap_home}
\end{subfigure}

\caption{
Extractability vs.\ Memorization (Magnitude) for \textit{LSTM-simple} on the \textit{Shanghai Telecom} dataset.
{Each violin shows the distribution of memorization magnitudes across extractability levels; width indicates density and central markers denote medians and quartiles statistics.
Lower extractability corresponds to higher vulnerability.
}}
\label{fig:rq4_memorization_vs_extractability}
\end{figure}

\subsection{Utility and Extractability} 
\label{subsec:rq4}


{We finally investigate how individual-level memorization impacts two downstream dimensions of model behavior: generalization performance (utility) and the risk of privacy leakage through extraction (extractability). In this work, we focus on extractability, where an adversary aims to recover sensitive trajectory information from partial observations. This setting captures a strong and practical form of privacy leakage, as it reflects the model’s ability to reconstruct previously unseen portions of a user’s behavior.}

{This perspective differs from membership inference, where the adversary is assumed to have access to a complete trajectory and seeks to determine whether it was part of the training set. While the attack objectives differ, both settings rely on a common underlying signal, namely, the model’s tendency to assign disproportionately high likelihood to memorized data. Our analysis therefore provides insight into privacy risks beyond a specific attack instantiation, revealing that memorization is closely entangled with how the model learns and what it can leak:}


\paragraph{Memorization harms utility.}
For each training trajectory, we compute the model's average top-1 accuracy on its corresponding reference set, capturing how well the model generalizes to unseen data. Figure~\ref{fig:rq4_memorization_vs_utility_hexbin} and Figure~\ref{fig:big_memo_utility_comparison} in the appendix reveal a consistent negative correlation between memorization (magnitude) and top-1 accuracy across datasets, indicating that higher memorization aligns with poorer generalization. 
Interestingly, we observe that memorization arises from two opposing mechanisms. In high-diversity settings like \emph{YJMob100K}, where 80\% of users have a diversity score above 0.95 (cf. Figure~\ref{fig:diversity_yjmob} {in appendix}), 
the model struggles to identify recurring patterns and defaults to memorizing trajectories, explaining both the low accuracy range (10–35\%) and the high magnitudes (0 to 7.5). By contrast, in more structured datasets, such as \emph{ShenzhenUrb}, memorization 
targets more users with low diversity, as shown in Figure~\ref{fig:rq2_heatmap_rg_diversity}. Here, the model captures highly learnable patterns, even when generalization would suffice. 

\medskip
\finding{Together, these findings highlight that memorization is not simply a byproduct of overfitting, but reflects both data complexity and the model’s inductive biases.}


\paragraph{Memorization amplifies extractability.}
Finally, we assess the risks posed by memorization through two realistic extraction proxies, each simulating a plausible inference scenario. Full details on attack design and metric formulation are provided in Appendix~\ref{appendix:extractability_attacks}.

\paragraph{(a) Prefix-based attack.}
For this attack, we assume the adversary has partial knowledge of a user’s trajectory and seeks to infer the remaining portion. We simulate three levels of attacker knowledge—prefixes of 6h, 12h, and 24h—and decode the rest of the 72h-long trajectory using a perplexity-weighted tree search. Extractability is measured as the logarithm of the number of greedy attempts required to retrieve the ground-truth suffix within the decoding tree. As shown in Figure~\ref{fig:rq4_violin_gap_prefix}, 
{trajectories with higher memorization magnitude 
tend to be easier to extract across prefix lengths, although the strength of this relationship varies across datasets (see Appendix \cref{fig:rq4_extractability_two_datasets}).}

\paragraph{(b) Anchor-based attack.}
In this attack, we align with the anchor-pair memorization setting, where an adversary knows one key location and attempts to infer its counterpart. In our simulation, we assume the attacker knows the user’s true home location and generates a synthetic trajectory initialized with stationary movement from 0–6 a.m. During the assumed commuting window (6–8 a.m.), we perform beam search to expand the trajectory. We then evaluate extractability over the subsequent work hours (9–17h) by ranking the true work location at each step and averaging these ranks across time. Figure~\ref{fig:rq4_violin_gap_home} shows that 
{users with higher anchor-pair memorization tend to be more vulnerable to work location inference in the \textit{ShanghaiTel} dataset, although this trend may be weaker across other datasets (Appendix \cref{fig:rq4_extractability_two_datasets}).} 


\medskip
\finding{We find that memorized samples act as inference shortcuts, surfacing more quickly during an attack. While these attacks serve as proxies, they approximate realistic adversarial capabilities and highlight concrete privacy risks.}

\section{Discussion and Implications}
\label{sec:discussion}
Our results demonstrate that memorization in mobility models is neither random nor negligible. 
{We discuss below key implications for privacy risk assessment and mitigation, as well as the framework scalability.}

\vspace{0.2em}
\emph{Memorization is predictable and user-specific.}
Our analyses show that memorization risk is not evenly distributed but disproportionately affects users with either highly regular or highly unique patterns. This observation unlocks new possibilities for mitigation. Rather than uniformly regularizing the dataset, one could adopt targeted training interventions, such as noise injection or dropout, applied more aggressively to memorization-prone users. 
Furthermore, model-level memorization was found to only partially reflect user-level retention in our experiments. This underscores the need for diverse audits using different metrics when deploying mobility models in sensitive contexts.

\vspace{0.2em}
\emph{Memorization covers behavioral patterns.}
Our findings show that memorization in mobility models is not limited to reproducing exact locations from the training set. Models also retain behavioral structure, such as routine fragments or anchor pairs. These forms of retention are not captured by traditional privacy evaluations that focus on exact memory leakage only. Our analysis therefore supports improved privacy assessments that account for behavioral memorization, where patterns indicative of personal habits are unintentionally stored in mobility models.

\vspace{0.2em}
\emph{Memorization exposes users to inference risks.}
We show that memorization correlates with how easily information can be extracted during inference, making our framework a useful early-warning tool for privacy vulnerabilities. By identifying trajectories that models treat as unusually likely, we can anticipate where attacks may succeed before they are carried out. 
{This connection is particularly clear when considering membership inference attacks. Such attacks rely on distinguishing whether a trajectory belongs to the training set by comparing its likelihood to that of non-training samples drawn from a similar distribution. In our framework, memorization explicitly captures this likelihood gap: each training trajectory is evaluated against a reference set of behaviorally similar but unseen trajectories, and high memorization corresponds to a significant deviation in likelihood. As a result, highly memorized trajectories are expected to be more easily distinguishable from non-training data, making them inherently more vulnerable to membership inference. This establishes a direct link between memorization and inference risk, based on the same statistical signal.}

\vspace{0.2em}
\emph{Cross-user memorization.}
{Our findings in Section~\ref{subsec:rq2} show that memorization increases with repetitiveness and stationarity, indicating that locations appearing more frequently within a trajectory are more likely to be retained by the model. This observation suggests a broader effect: locations that are frequently observed—either within a single trajectory or across multiple users—are inherently easier for the model to memorize. For instance, popular places (e.g., transport hubs or event venues) may be learned more strongly due to their repeated occurrence in the data.
However, such cross-user memorization does not necessarily imply the same level of privacy risk as user-specific memorization. While frequently visited locations may be memorized at the population level, privacy risks arise when these locations can be associated with an individual’s trajectory or routine. 
Overall, analyzing cross-user location memorization, and disentangling global popularity from user-specific patterns, is a promising direction for future work.}

\vspace{0.2em}
\emph{Scalability of the framework.}
{The proposed framework evaluates memorization using a representative subset of trajectories rather than requiring exhaustive processing of the full dataset. In our setup, $2{,}000$ medoid trajectories are selected via clustering, capturing diverse mobility behaviors and enabling an efficient approximation of memorization at the dataset level.}
{
The main computational cost arises from the reference set construction phase, which includes trajectory abstraction, clustering, and dataset generation. Importantly, this step is performed only once per dataset, and the resulting reference sets can be reused across different models. Consequently, reference set construction cost is primarily driven by dataset size and abstraction complexity (see Appendix Table~\ref{tab:runtime_reference}).}
{
In contrast, memorization assessment is lightweight (see Appendix Table~\ref{tab:runtime_assessment}). While perplexity computation depends on the underlying model and dominates the evaluation runtime, the memorization metrics themselves require only a few seconds across all datasets and models. This makes the proposed framework practical for large-scale evaluation, as the additional overhead beyond standard model inference is negligible.}

\section{Conclusion and Perspective}
\label{sec:conclusion}
Unintended memorization is a pervasive problem in mobility prediction models. In our experiments, we show that memorized data is not a simple side effect of overfitting, but arises depending on user behavior and model design. By examining three memorization types (location level, anchor pair, and segment level memorization), we demonstrate that the models unintentionally store sensitive parts of people's movements in their parameters, ranging from personal locations to fragments of daily routes. Unfortunately, these memorized patterns are easy to extract, revealing a concrete privacy risk in current prediction systems.

Our framework provides the first systematic means for assessing this risk, yet it should not be interpreted as the ultimate approach to auditing privacy in mobility models. The abstractions we study and the clustering method we use represent only one possible way to measure memorization, given the varying granularity of privacy leakage. That is, they constitute a best-effort design for quantifying this behavior and should thus rather serve as a template for similar analyses in future research.

Overall, our findings echo a broader privacy problem in generative modeling. Whenever sensitive data are used to train generative models, there is a risk that parts of the training data are memorized and can be exposed by adversaries. This problem is inherent to generative learning, which must strike a balance between capturing general patterns and relying on specific examples in the training data. The resulting risk is especially pronounced for mobility models, which are widely used in practice yet rely on highly sensitive location data. We hope that our framework provides a new practical tool for assessing this risk in deployed systems.


\section*{Acknowledgments}
Generative AI tools (ChatGPT by OpenAI) were used solely for minor editorial
assistance, including improving text flow and correcting typographical and
grammatical issues.

This work was supported by the European Research Council (ERC) under the Consolidator Grant MALFOY (Grant Agreement No. 101043410) and by the German Federal Ministry of Research, Technology and Space (BMFTR) under the grant AIgenCY (16KIS2012). The work of Anne Josiane Kouam was supported through the Mob Sci-Dat Factory project (ANR-23-PEMO-0004) under the France 2030 program.

\begin{acks}
Generative AI tools (ChatGPT by OpenAI) were used solely for minor editorial
assistance, including improving text flow and correcting typographical and
grammatical issues.
\end{acks}

\bibliographystyle{ACM-Reference-Format}
\balance
\bibliography{references}

\appendix

\section{Open Science}
\label{app:open_science}

To support the results presented in this paper and enable reproducibility, we
release the full experimental artifact used in our evaluation. The artifact
includes the data preprocessing pipeline, reference-set construction procedures,
model training and evaluation code, memorization auditing scripts, and result
analysis tools.

\paragraph{Datasets.}
Our experiments rely exclusively on publicly available mobility datasets:
\begin{itemize}[leftmargin=*]
    \item \textbf{Shanghai Telecom}.  
    Available from Kaggle at \cite{shanghai_kaggle}.
    The dataset consists of 12 files (XLSX format) containing anonymized cellular
    activity records, all of which are used in our experiments.

    \item \textbf{YJMob100K}.  
    Available from Zenodo at \cite{yjmob100k}.
    We use the file \texttt{yjmob100k-dataset1.csv.gz}, which corresponds to the
    standard mobility period.

    \item \textbf{Shenzhen Urban}.  
    Publicly available at \cite{ShenzhenUrbanData}.
    From the datasets released under \emph{Urban Data Release v2}, we use only the
    Call Detail Record data. Other data modalities provided on the same page
    (e.g., smart card, taxi GPS, bus GPS) are not used in this work.
\end{itemize}

The raw datasets are not redistributed. The artifact provides detailed
instructions to download the data from their official sources and to reproduce
the preprocessing steps described in the paper.

\paragraph{Repository access.}
The complete artifact is available at the following repository:
\begin{center}
\href{https://gitlab.inria.fr/mobleak/mobleak-predictive}{\texttt{gitlab.inria.fr/mobleak/mobleak-predictive}}
\end{center}

\paragraph{Released artifact.}
The released artifact contains:
(i) preprocessing notebooks implementing all data transformations and trajectory
construction steps;
(ii) scripts to generate training sets and reference sets for the three
memorization abstractions studied (location-level, anchor-pair, and segment-level);
(iii) implementations or adapted wrappers for all evaluated mobility prediction
models;
(iv) an automated pipeline to compute trajectory-level memorization metrics across
models and datasets; and
(v) analysis scripts to reproduce the figures and empirical findings reported in
the paper.


\begin{algorithm}[t]
\caption{Cluster Enrichment for Reference Set Balancing}
\label{alg:enrichment}
\begin{algorithmic}[1]
\Require Trajectory records with cluster label and abstraction vector; cluster medoids; target minimum reference size $k_{\min}$; distance threshold $\tau$
\State {\textit{$k_{\min}$ defines a target minimum size; reaching it depends on $\tau$ and the data distribution}}
\State Build mapping from each cluster to its member trajectories
\State Build mapping from each medoid to its abstraction vector
\State Compute inter-cluster distances between medoid abstractions
\For{each cluster $C$}
    \State Select medoid trajectory $T_u$ in $C$ as training sample
    \State Initialize reference set $\mathcal{R}_C \gets C \setminus \{T_u\}$
    \If{$|\mathcal{R}_C| < k_{\min}$}
        \State Order other clusters by increasing medoid distance to $C$
        \For{each neighboring cluster $C'$ in this order}
            \For{each trajectory $T' \in C'$ sorted by distance to $T_u$}
                \If{$d(T', T_u) \le \tau$}
                    \State Add $T'$ to $\mathcal{R}_C$
                \EndIf
                \If{$|\mathcal{R}_C| \ge k_{\min}$}
                    \State \textbf{break} out of both loops
                \EndIf
            \EndFor
        \EndFor
    \EndIf
    \State Set $\tilde{\mathcal{R}} (T_u) \gets \mathcal{R}_C$
\EndFor
\State \Return training set $\{T_u\}$ and reference sets $\{\tilde{\mathcal{R}} (T_u)\}$
\end{algorithmic}
\end{algorithm}

\section{Mobility Metrics and Profiles}
\label{appendix:mobility_metrics}

The following mobility metrics are widely used in the literature to capture both spatial and structural properties of individual behavior. Below we provide the definitions and formulas used in our study.

\vspace{1em}
\noindent\textbf{Repetitiveness.} \quad
Repetitiveness measures how often a user returns to previously visited places \cite{mucceli:2016}. It is defined as the fraction of visited locations that revisit an already-seen place:
\[
\text{rep} = 1 - \frac{n_{\text{unique}}}{n},
\]
where \(n\) is the trajectory length and \(n_{\text{unique}}\) counts distinct locations. High values indicate routine-heavy movement.

\vspace{0.8em}
\noindent\textbf{Stationarity.} \quad
Stationarity captures temporal persistence in user movement \cite{mucceli:2016}. It corresponds to the proportion of consecutive timestamps at which the user remains in the same location:
\[
\text{sta} = \frac{1}{n-1}\sum_{i=1}^{n-1} \mathbb{1}[l_i = l_{i+1}].
\]
Higher stationarity reflects long stays or sedentary behavior.

\begin{figure}[t]
\centering
\begin{subfigure}{0.32\linewidth}
    \centering
    \includegraphics[width=\linewidth]{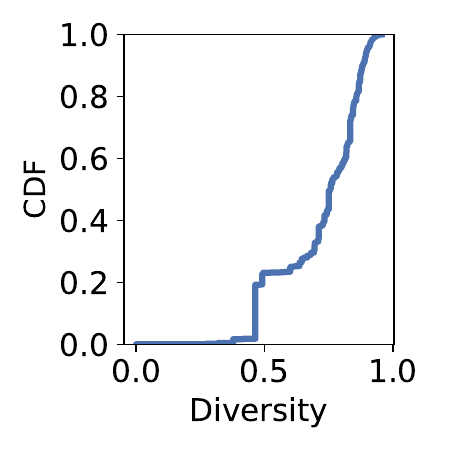}
    \caption{ShenzhenUrban}
    \label{fig:diversity_shenzhen}
\end{subfigure}
\hfill
\begin{subfigure}{0.32\linewidth}
    \centering
    \includegraphics[width=\linewidth]{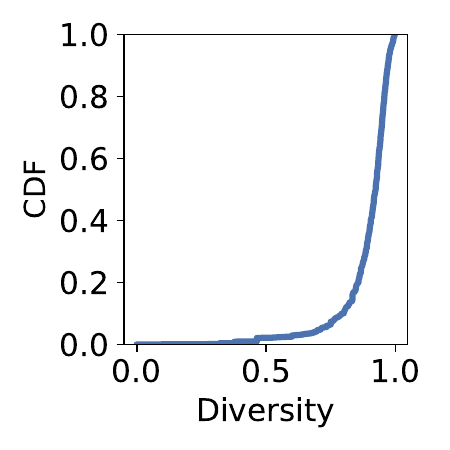}
    \caption{ShanghaiTelecom}
    \label{fig:diversity_shanghai}
\end{subfigure}
\hfill
\begin{subfigure}{0.32\linewidth}
    \centering
    \includegraphics[width=\linewidth]{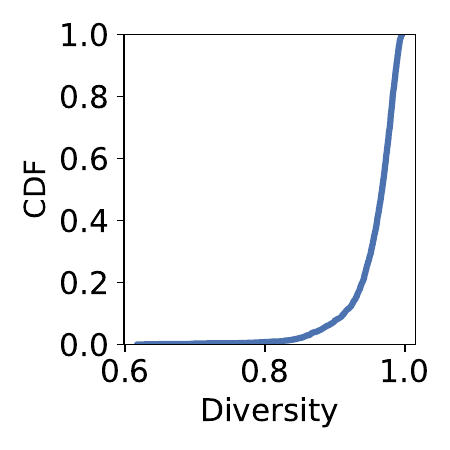}
    \caption{YJMob100K}
    \label{fig:diversity_yjmob}
\end{subfigure}
\caption{Distribution of mobility diversity across datasets.}
\label{fig:appendix_diversity_distribution}
\end{figure}

\vspace{0.8em}
\noindent\textbf{Diversity.} \quad
Diversity quantifies the structural richness of a trajectory by counting the number of distinct sub-trajectories it contains \cite{teixeira:hal-03360537}. Formally, for a sequence partitioned into \(m\) sub-trajectories \(\{\tau_1,\dots,\tau_m\}\),
\[
\text{div} = |\{\tau_i \;|\; \tau_i \text{ is unique}\}|.
\]
Higher diversity indicates more variable mobility.
Figure~\ref{fig:appendix_diversity_distribution} shows the distribution of diversity across datasets, indicating that users in \emph{YJMob100K} exhibit higher spatial entropy in their movements, followed by users in \emph{ShanghaiTelecom} and \emph{ShenzhenUrban}.

\vspace{0.8em}
\noindent\textbf{Radius of Gyration.} \quad
The radius of gyration measures the spatial extent of user movement \cite{Gonzalez:2008}. It computes the average distance of visited locations from the trajectory’s center of mass:
\[
r_g = \sqrt{\frac{1}{n}\sum_{k=1}^n \lVert l_k - \bar{l} \rVert^2},
\qquad
\bar{l} = \frac{1}{n}\sum_{k=1}^n l_k.
\]
Large \(r_g\) reflects broad geographic mobility, while small \(r_g\) signals localized behavior.

\vspace{1em}
\noindent\textbf{Mobility Profiles.} \quad 
Human mobility behaviors vary significantly across individuals. To account for this heterogeneity, we follow the categorization proposed by  \citet{Amichi:2020}, who identified three core mobility profiles based on visitation regularity and behavioral dynamics: (i) \textit{Scouters}, who frequently explore new locations with high spatial variability; (ii) \textit{Routiners}, who exhibit repetitive and localized movement patterns; and (iii) \textit{Regulars}, who balance exploration with routine. These profiles are derived by clustering users according to trajectory-level return and exploration behaviors (i.e., number of successive returns and explorations), and serve as interpretable proxies for user-level predictability. 

\begin{table*}[t]
\centering
\small
\caption{Model hyperparameters and training settings. Embedding and hidden sizes are given in dimensions. "Reg." denotes L2 or weight decay regularization.}
\label{tab:model-hparams}
\begin{tabular}{lcccccccc}
\toprule
                     & \multicolumn{3}{c}{Embedding Size} &        &      &      &       &        \\
\cmidrule(lr){2-4}
Model                & Location    & Time      & User     & (Hidden) Units & LR                   & Reg.      & Batch & Epochs \\
\midrule
Markov (2nd)         & --          & --        & --       & 2     & --                   & --        & --    & --    \\
LSTM-simple          & 500         & 10        & 40       & 500    & $\tfrac{1}{3}\times10^{-4}$ & $10^{-5}$  & 64    & 40    \\
LSTM-long            & 500         & 10        & 40       & 500    & $\tfrac{1}{3}\times10^{-4}$ & $10^{-5}$  & 64    & 40    \\
DM-locallong         & 500         & 10        & 40       & 500    & $\tfrac{1}{3}\times10^{-4}$ & $10^{-5}$  & 64    & 40    \\
DM-avglonguser       & 500         & 10        & 40       & 500    & $\tfrac{1}{3}\times10^{-4}$ & $10^{-5}$  & 64    & 40    \\
LSTPM                & 500         & 10        & --       & 500    & $10^{-4}$            & $10^{-6}$ & 256   & 100   \\
Graph-Flashback      & 10          & --        & 10       & 10     & $10^{-2}$            & $0$       & 200   & 100   \\
\bottomrule
\end{tabular}
\end{table*}

\begin{figure}
    \centering
    \includegraphics[width=\linewidth]{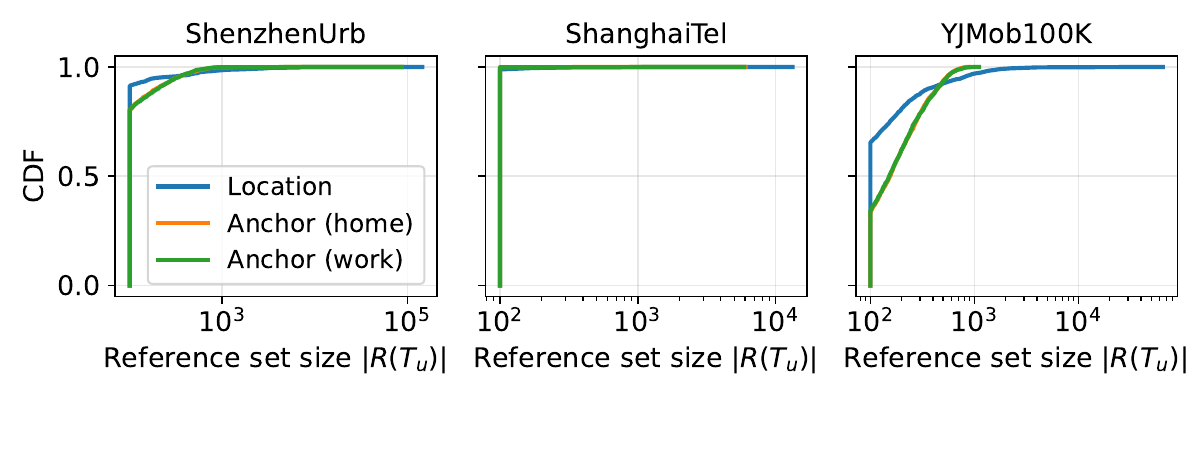}
    \caption{CDF of reference-set sizes across datasets.}
    \label{fig:refset_sizes}
\end{figure}

\begin{figure}
    \centering
    \includegraphics[width=\linewidth]{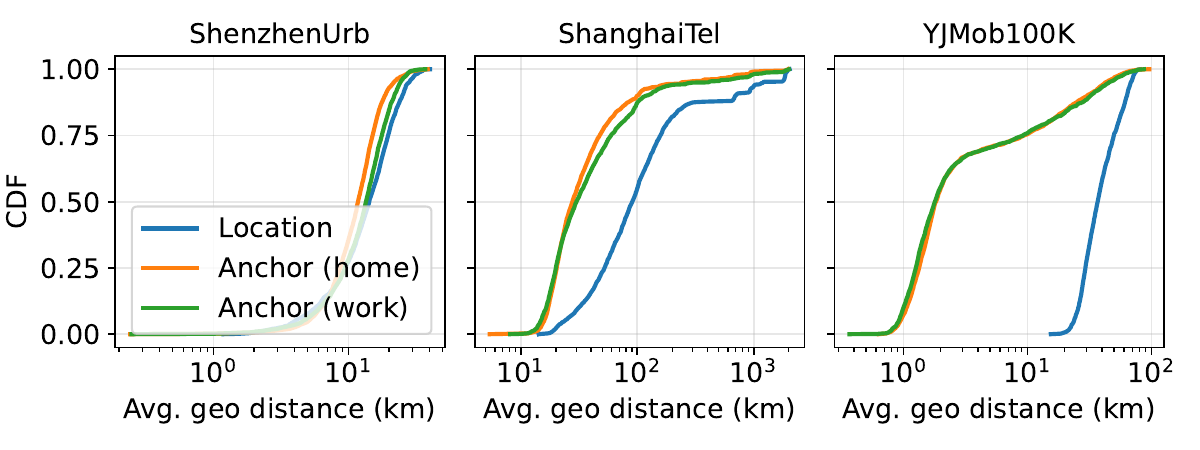}
    \caption{
        CDF of average spatial dispersion within reference sets, by dataset and memorization type. For each training trajectory, we compute the mean geographic distance (in km) between its origin and the origins of all reference trajectories in its cluster. 
        For \texttt{YJMob100K}, distances are computed using the dataset’s grid system (500\,m per cell). Larger values indicate that reference sets contain spatially-diverse but plausible alternatives rather than near-duplicates.
    }
    \label{fig:refset_geo_dispersion}
\end{figure}

\begin{figure*}[ht]
\centering
\begin{subfigure}{0.32\textwidth}
    \centering
    \includegraphics[width=\linewidth]{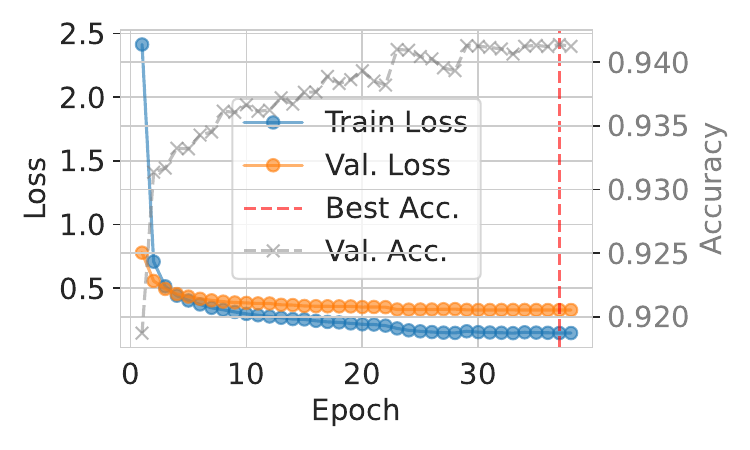}
    \caption{Shenzhen Urban}
\end{subfigure}
\hfill
\begin{subfigure}{0.32\textwidth}
    \centering
    \includegraphics[width=\linewidth]{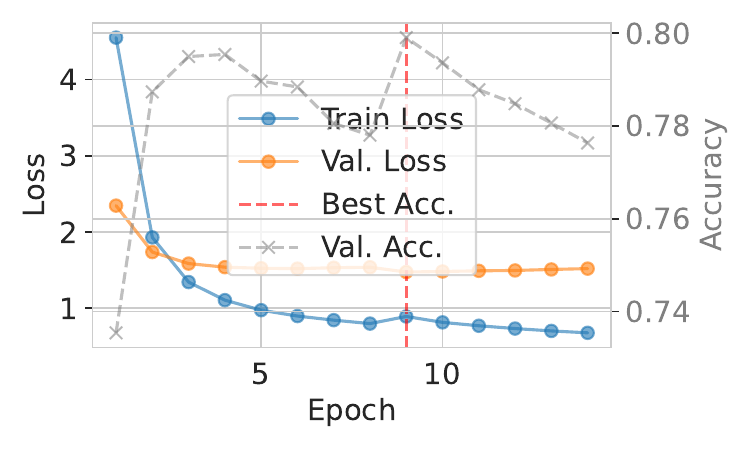}
    \caption{Shanghai Telecom}
\end{subfigure}
\hfill
\begin{subfigure}{0.32\textwidth}
    \centering
    \includegraphics[width=\linewidth]{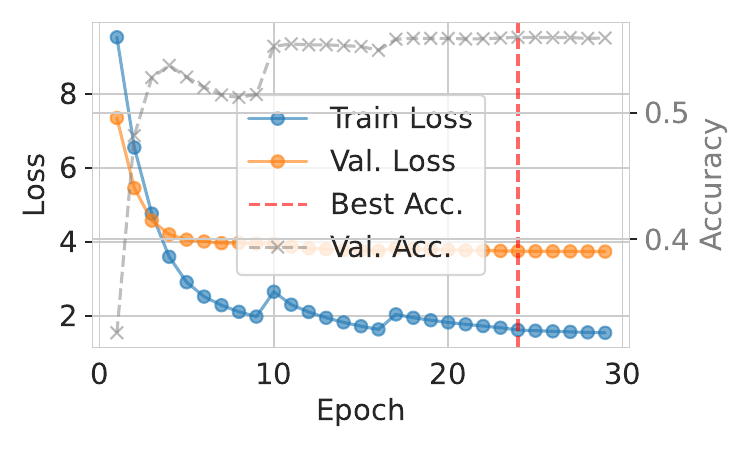}
    \caption{YJMob100K}
\end{subfigure}

\caption{Training dynamics for the LSTM-simple model across all datasets: Training and validation loss decrease steadily and do not diverge, indicating an underfitted but stable regime suitable for analyzing memorization behavior.}
\label{fig:app_training_dynamics}
\end{figure*}

\begin{figure*}[htbp]
\centering
\begin{subfigure}{0.299\textwidth}
    \centering
    \includegraphics[width=\linewidth]{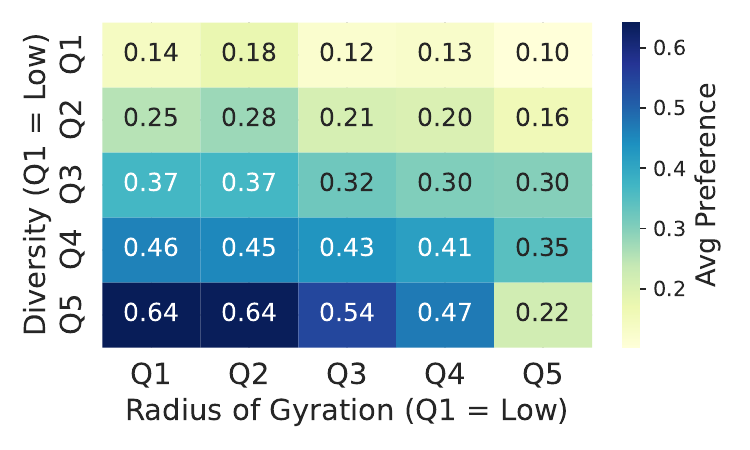}
    \caption{Preference (div vs. rg)}
    \label{fig:}
\end{subfigure}
\hfill
\begin{subfigure}{0.25\textwidth}
    \centering
    \includegraphics[width=\linewidth]{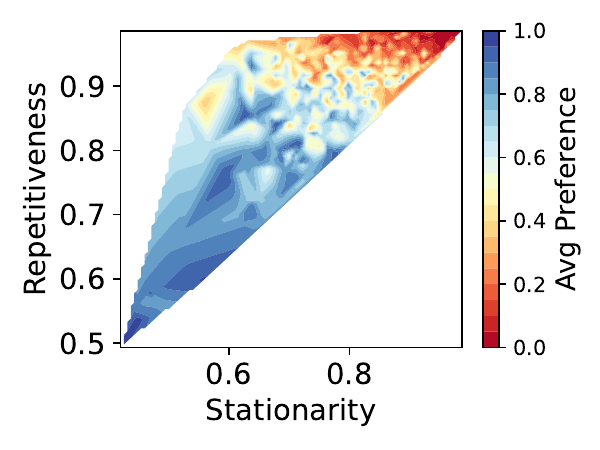}
    \caption{Preference (sta vs. rep)}
    \label{fig:}
\end{subfigure}
\hfill
\begin{subfigure}{0.22\textwidth}
    \centering
    \includegraphics[width=\linewidth]{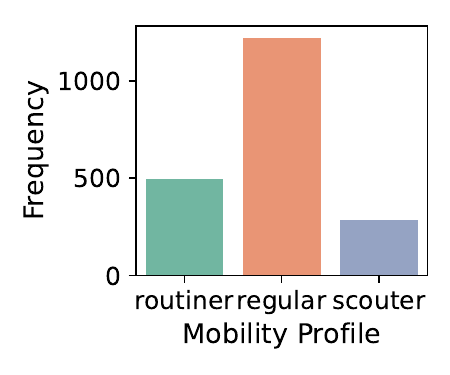}
    \caption{Mobility profiles.}
    \label{fig:}
\end{subfigure}
\hfill
\begin{subfigure}{0.19\textwidth}
    \centering
    \includegraphics[width=\linewidth]{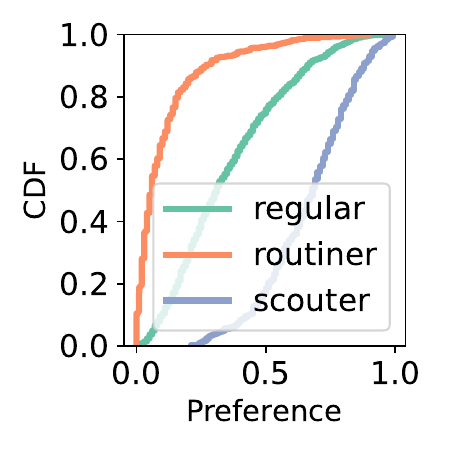}
    \caption{Preference (profiles)}
    \label{fig:}
\end{subfigure}

\begin{subfigure}{0.299\textwidth}
    \centering
    \includegraphics[width=\linewidth]{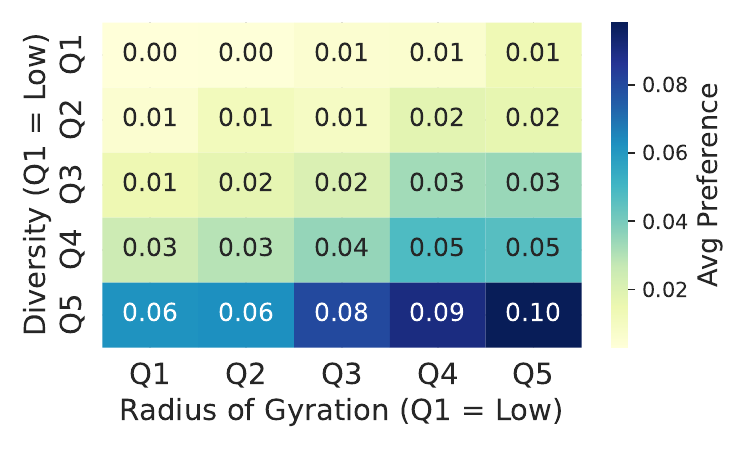}
    \caption{Preference (div vs. rg)}
    \label{fig:}
\end{subfigure}
\hfill
\begin{subfigure}{0.25\textwidth}
    \centering
    \includegraphics[width=\linewidth]{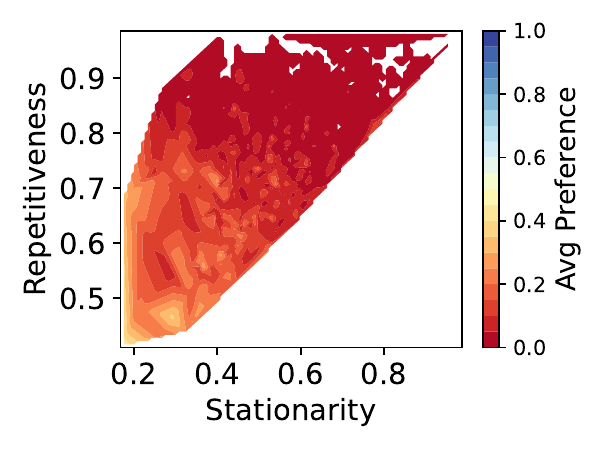}
    \caption{Preference (sta vs. rep)}
    \label{fig:}
\end{subfigure}
\hfill
\begin{subfigure}{0.22\textwidth}
    \centering
    \includegraphics[width=\linewidth]{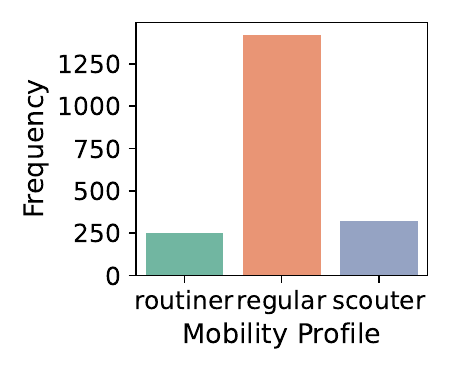}
    \caption{Mobility profiles.}
    \label{fig:}
\end{subfigure}
\hfill
\begin{subfigure}{0.19\textwidth}
    \centering
    \includegraphics[width=\linewidth]{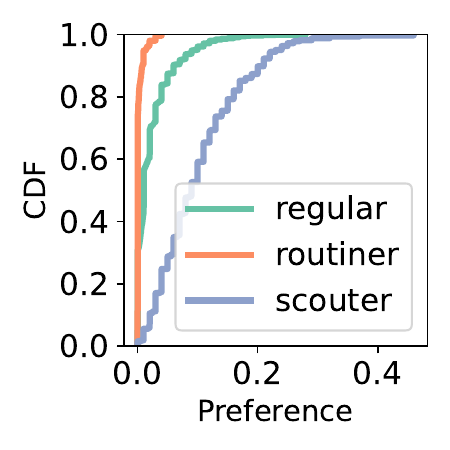}
    \caption{Preference (profiles).}
    \label{fig:}
\end{subfigure}
\caption{Mobility characteristics and memorization patterns: Exposure preference for location memorization on the model LSTM-simple trained on the Shanghai Telecom dataset (a)-(d) and YJMob100k  dataset (e)-(g).}
\label{fig:rq2_memorization_vs_mobility_yjmob}
\end{figure*}

{\section{Additional Results and Analysis}}
\label{appendix:results}

{\subsection{Memorization and Mobility Characteristics}}
\label{appendix:rq2_details}

{We provide additional graphs supporting the analysis of the relationship between memorization and mobility characteristics. Figure~\ref{fig:rq2_memorization_vs_mobility_yjmob} extends the results presented in Figure~\ref{fig:rq2_memorization_vs_mobility} (main paper, \textit{ShenzhenUrb}) by reporting the same analysis on \textit{ShanghaiTel} and \textit{YJMob100K}.}

{Compared to the \textit{ShenzhenUrb} results, the appendix figures exhibit noticeable differences in value ranges and visual distributions across datasets. On a common scale, \textit{YJMob100K} exhibits lower preference values (indicating higher memorization), whereas \textit{ShanghaiTel} concentrates in higher preference values (indicating lower memorization), with \textit{ShenzhenUrb} lying in between. 
These shifts directly impact the visual patterns: in \textit{ShanghaiTel}, the contour plot is dominated by high preference values (blue regions), with only limited areas of lower preference, whereas in \textit{YJMob100K}, the distribution shifts toward lower preference values, resulting in contour plots largely dominated by red and yellow regions (Figures~\ref{fig:rq2_memorization_vs_mobility_yjmob}b,f).  Overall, these differences reflect changes in memorization intensity across datasets rather than in the underlying relationships between mobility features and memorization.}

{Despite these variations in magnitude, the relationships between mobility features and memorization remain stable across datasets. Memorization consistently increases with greater \textit{stationarity} and \textit{repetitiveness}, indicating that users with more persistent movement patterns (routiners) are more likely to be memorized. In contrast, higher \textit{diversity} is associated with lower memorization, reflecting the difficulty of models to capture highly variable behaviors.}

{The effect of the radius of gyration is weaker and less consistent across datasets. This suggests that spatial extent alone is not a primary driver of memorization, in contrast to temporal regularity features such as stationarity and repetitiveness, which exhibit a clearer and more stable influence.}

{\paragraph{Summary.}
Overall, the results confirm that while the absolute level of memorization varies with the characteristics of the datasets, the structural relationships between mobility behavior and memorization remain robust.}

{\subsection{Memorization Across Abstractions, Models, and Temporal Patterns}}
\label{appendix:rq3_details}

\begin{figure*}[htbp]
\centering
\begin{subfigure}{0.23\textwidth}
    \centering
    
    \includegraphics[width=\linewidth]{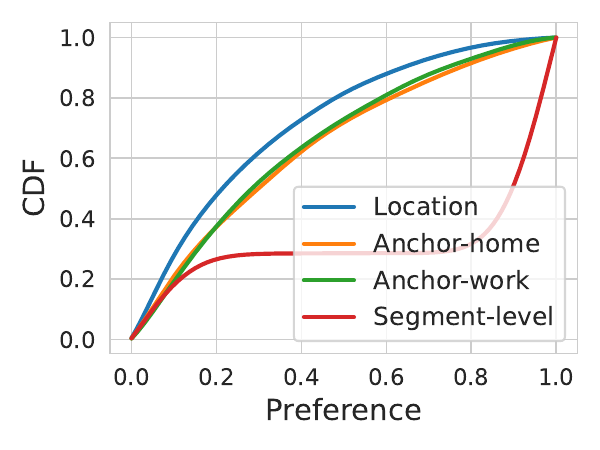}
    \caption{Preference distribution across memorization risks.}
    \label{fig:}
\end{subfigure}
\hfill
\begin{subfigure}{0.23\textwidth}
    \centering
    \includegraphics[width=\linewidth]{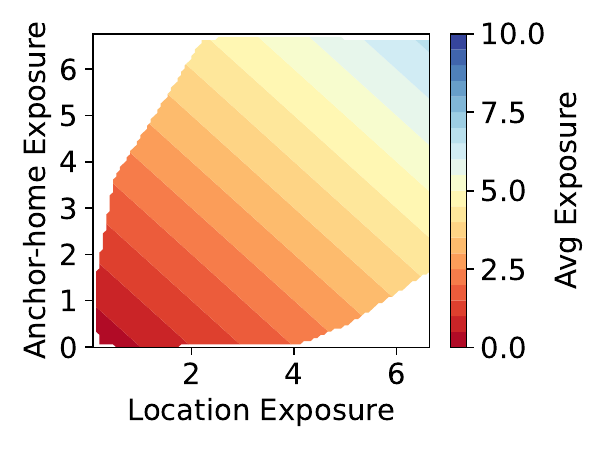}
    \caption{Exposure comparison: \textit{Location} vs. \textit{Anchor-home}.}
    \label{fig:}
\end{subfigure}
\hfill
\begin{subfigure}{0.23\textwidth}
    \centering
    \includegraphics[width=\linewidth]{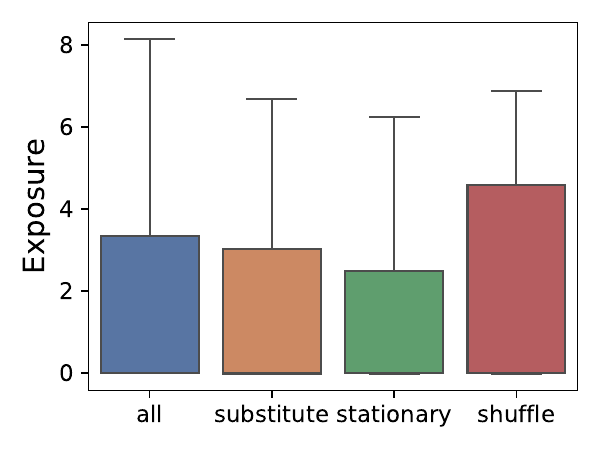}
    \caption{Exposure variation across pertubation modes (\textit{Segment-level}).}
    \label{fig:}
\end{subfigure}
\hfill
\begin{subfigure}{0.23\textwidth}
    \centering
    \includegraphics[width=\linewidth]{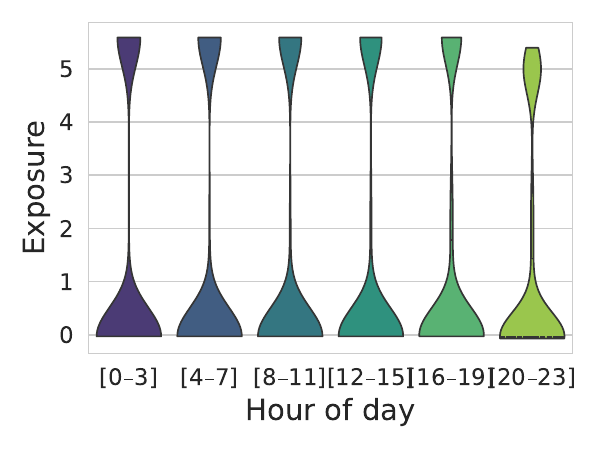}
    \caption{Exposure by time-of-day segments (\textit{Segment-level}).}
    \label{fig:}
\end{subfigure}



\begin{subfigure}{0.23\textwidth}
    \centering
    
    \includegraphics[width=\linewidth]{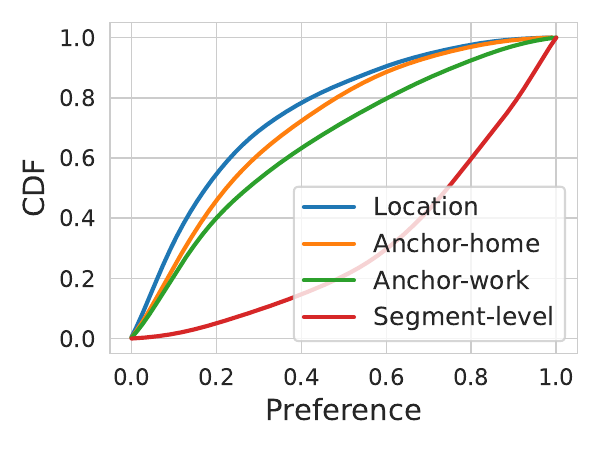}
    \caption{Preference distribution across memorization risks.}
    \label{fig:}
\end{subfigure}
\hfill
\begin{subfigure}{0.23\textwidth}
    \centering
    \includegraphics[width=\linewidth]{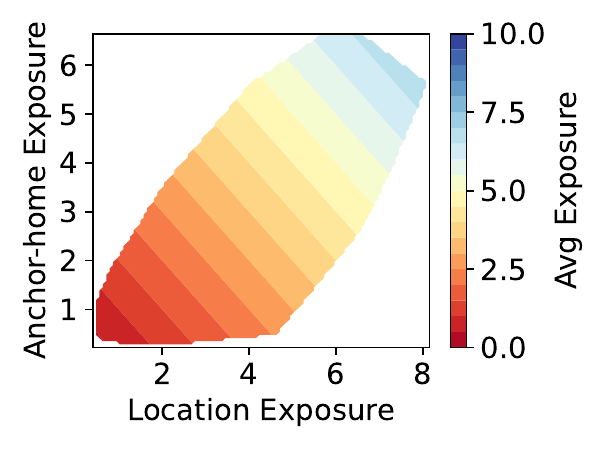}
    \caption{Exposure comparison: \textit{Location} vs. \textit{Anchor-home}.}
    \label{fig:}
\end{subfigure}
\hfill
\begin{subfigure}{0.23\textwidth}
    \centering
    \includegraphics[width=\linewidth]{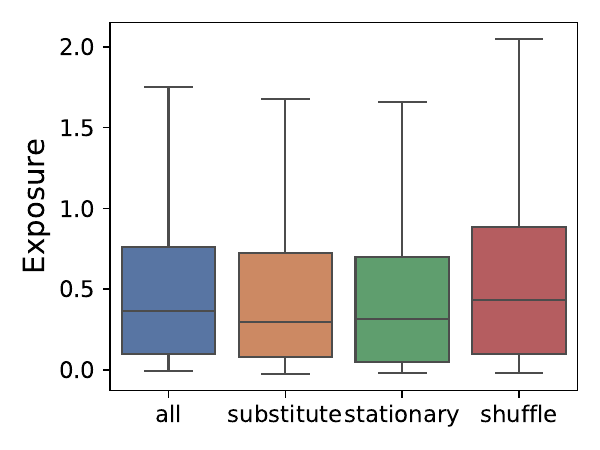}
    \caption{Exposure variation across pertubation modes (\textit{Segment-level}).}
    \label{fig:}
\end{subfigure}
\hfill
\begin{subfigure}{0.23\textwidth}
    \centering
    \includegraphics[width=\linewidth]{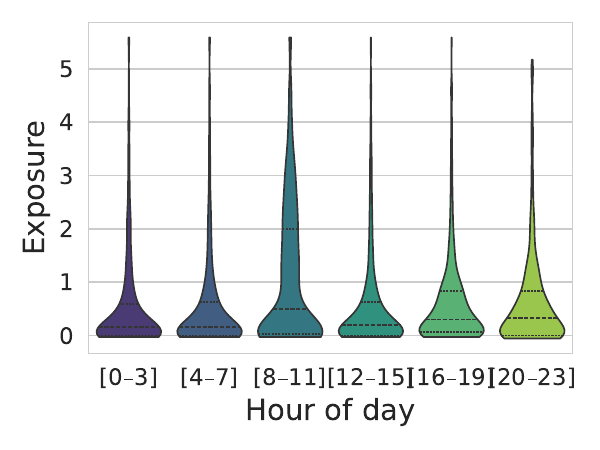}
    \caption{Exposure by time-of-day segments (\textit{Segment-level}).}
    \label{fig:}
\end{subfigure}


\begin{subfigure}{0.23\textwidth}
    \centering
    
    \includegraphics[width=\linewidth]{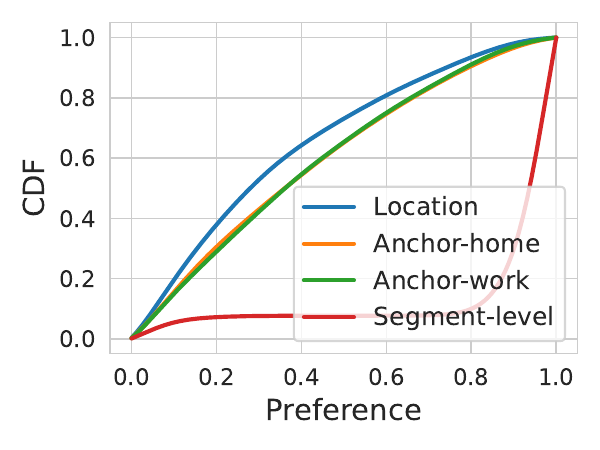}
    \caption{Preference distribution across memorization risks.}
    \label{fig:}
\end{subfigure}
\hfill
\begin{subfigure}{0.23\textwidth}
    \centering
    \includegraphics[width=\linewidth]{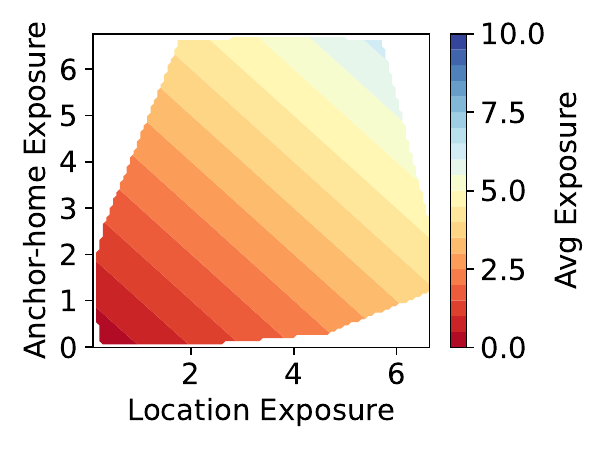}
    \caption{Exposure comparison: \textit{Location} vs. \textit{Anchor-home}.}
    \label{fig:}
\end{subfigure}
\hfill
\begin{subfigure}{0.23\textwidth}
    \centering
    \includegraphics[width=\linewidth]{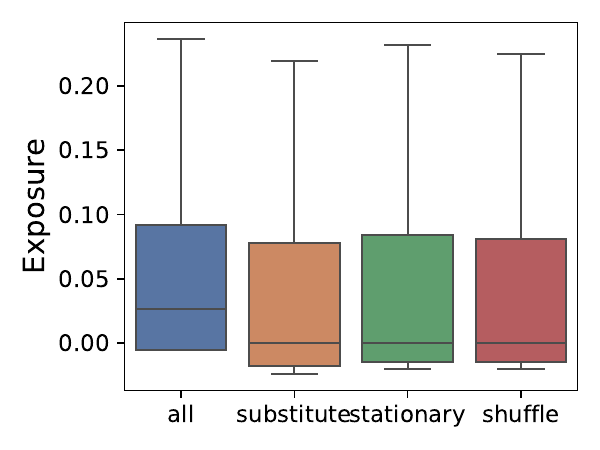}
    \caption{Exposure variation across pertubation modes (\textit{Segment-level}).}
    \label{fig:}
\end{subfigure}
\hfill
\begin{subfigure}{0.23\textwidth}
    \centering
    \includegraphics[width=\linewidth]{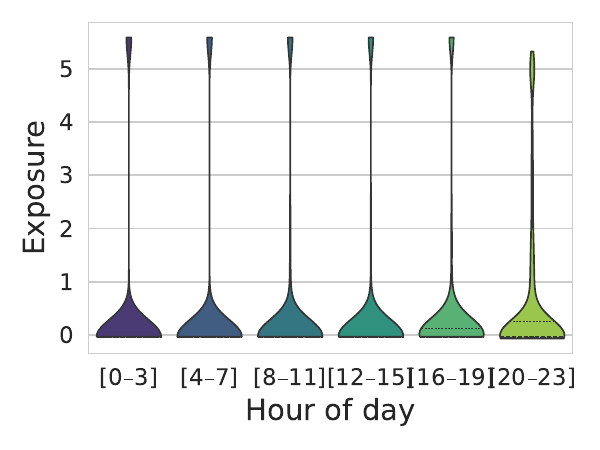}
    \caption{Exposure by time-of-day segments (\textit{Segment-level}).}
    \label{fig:}
\end{subfigure}



\begin{subfigure}{0.23\textwidth}
    \centering
    
    \includegraphics[width=\linewidth]{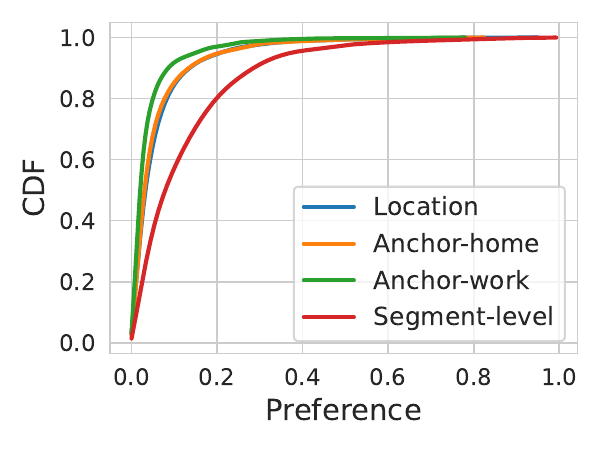}
    \caption{Preference distribution across memorization risks.}
    \label{fig:}
\end{subfigure}
\hfill
\begin{subfigure}{0.23\textwidth}
    \centering
    \includegraphics[width=\linewidth]{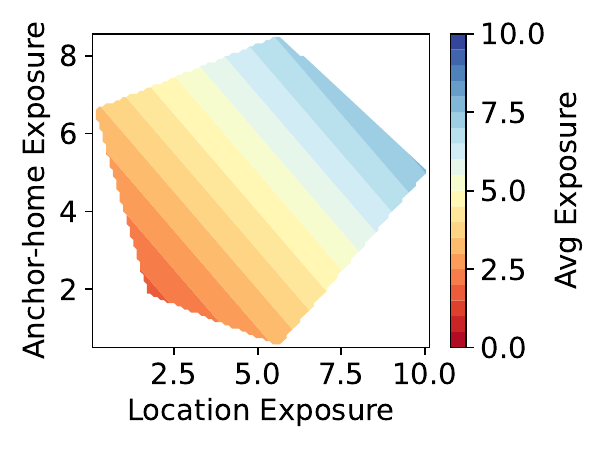}
    \caption{Exposure comparison: \textit{Location} vs. \textit{Anchor-home}.}
    \label{fig:}
\end{subfigure}
\hfill
\begin{subfigure}{0.23\textwidth}
    \centering
    \includegraphics[width=\linewidth]{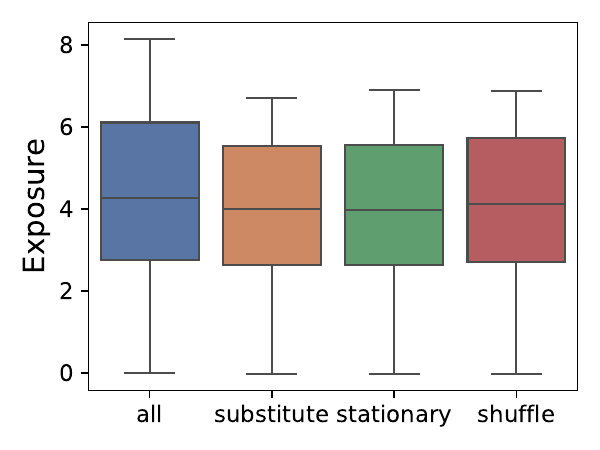}
    \caption{Exposure variation across pertubation modes (\textit{Segment-level}).}
    \label{fig:}
\end{subfigure}
\hfill
\begin{subfigure}{0.23\textwidth}
    \centering
    \includegraphics[width=\linewidth]{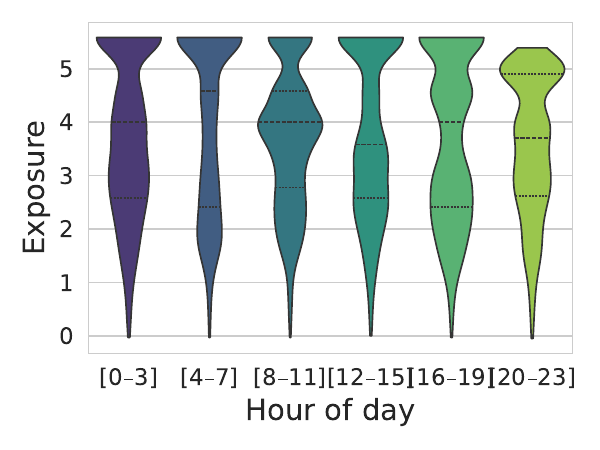}
    \caption{Exposure by time-of-day segments (\textit{Segment-level}).}
    \label{fig:}
\end{subfigure}

\caption{Memorization behavior across risks, transformations, and sub-trajectories of the following models trainined on Shenzhen Urban dataset: (a)-(d) \textit{Markov}, (e)-(h) \textit{DM-avglonguser}, (i)-(l) \textit{LSTPM}, and (m)-(p) \emph{Graph-flashback}. 
}
\label{fig:rq3_memorization_patterns_appendix}
\end{figure*}

\begin{figure*}[htbp]
\centering

\begin{subfigure}{0.19\textwidth}
    \centering
    \includegraphics[width=\linewidth]{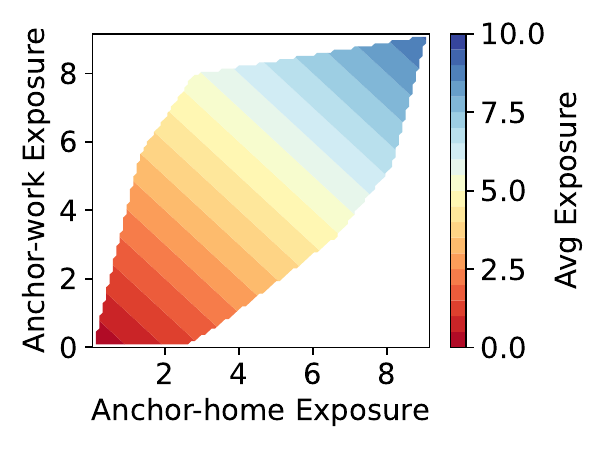}
    \caption{{LSTM-simple}}
    \label{fig:rq3_contour_home_vs_work_lstm_simple}
\end{subfigure}
\hfill
\begin{subfigure}{0.19\textwidth}
    \centering
    \includegraphics[width=\linewidth]{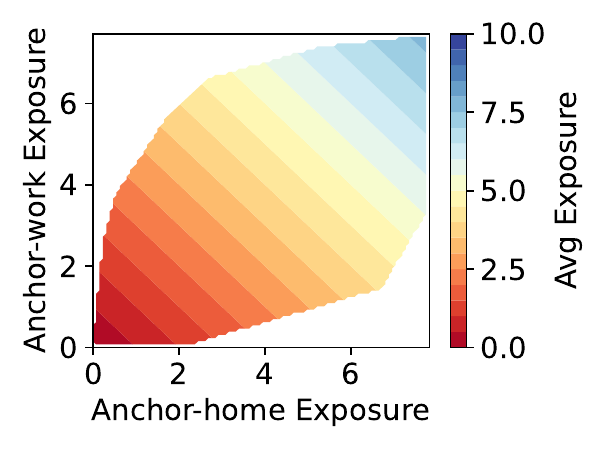}
    \caption{{Markov}}
    \label{fig:rq3_contour_home_vs_work_lstm_long}
\end{subfigure}
\hfill
\begin{subfigure}{0.19\textwidth}
    \centering
    \includegraphics[width=\linewidth]{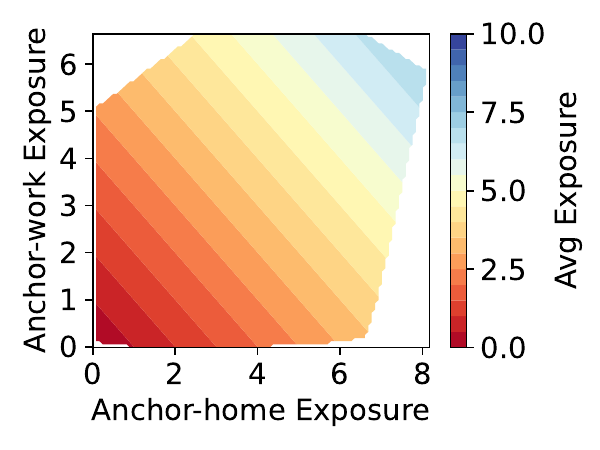}
    \caption{{DM-avglonguser}}
    \label{fig:rq3_contour_home_vs_work_dm_avglonguser}
\end{subfigure}
\hfill
\begin{subfigure}{0.19\textwidth}
    \centering
    \includegraphics[width=\linewidth]{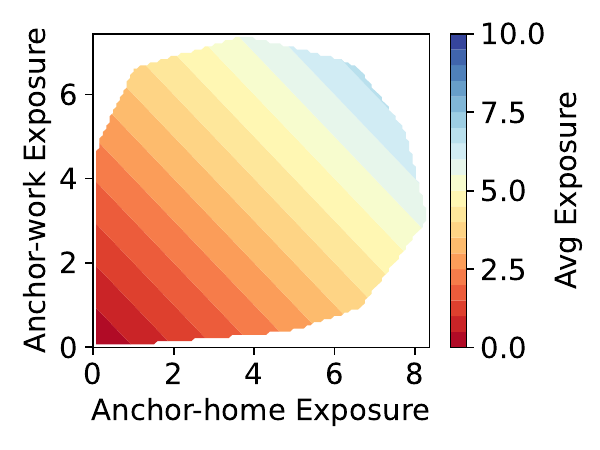}
    \caption{{LSTPM}}
    \label{fig:rq3_contour_home_vs_work_dm_locallong}
\end{subfigure}
\hfill
\begin{subfigure}{0.19\textwidth}
    \centering
    \includegraphics[width=\linewidth]{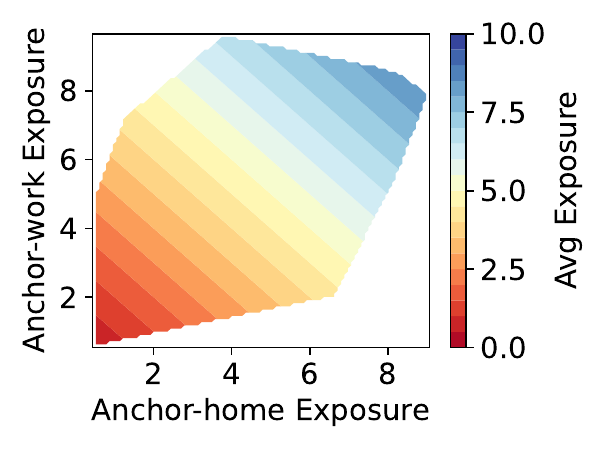}
    \caption{{Graph-Flashback}}
    \label{fig:rq3_contour_home_vs_work_graphflashback}
\end{subfigure}

\caption{Comparison of memorization exposure between \textit{Anchor-home} and \textit{Anchor-work} abstractions for different model architectures trained on \texttt{ShenzhenUrban}.}
\label{fig:rq3_contour_home_vs_work_all_models}
\end{figure*}

{
We provide additional analyses to the results presented in Figure~\ref{fig:rq3_memorization_patterns} (main paper), which focuses on the \textit{LSTM-simple} model. Figures~\ref{fig:rq3_contour_home_vs_work_all_models} and~\ref{fig:rq3_memorization_patterns_appendix} extend this analysis to multiple model architectures, for a more comprehensive assessment of memorization behavior.}

{
\paragraph{Memorization across abstractions.}
Consistent with the main paper, location-based memorization generally exhibits the lowest preference values (i.e., highest memorization), followed by anchor-based abstractions, while segment-level patterns show substantially higher preference values. However, Figures~\ref{fig:rq3_memorization_patterns_appendix}a,e,i,m reveal that the strength of this separation varies across models. While neural architectures such as \textit{DeepMove} and \textit{LSTPM} preserve a clear ordering between abstraction levels, simpler models (e.g., \textit{Markov}) and lower-capacity architectures (e.g., \textit{Graph-Flashback}) exhibit more compressed distributions, reducing the contrast between abstractions. Hence, abstraction-level memorization is not uniform across models, but depends on their ability to capture long-term structure.}

{
\paragraph{Cross-abstraction relationships.}
Figure~\ref{fig:rq3_contour_home_vs_work_all_models} shows that the positive relationship between location-based and anchor-based memorization observed in the main paper remains visible across models. Users with higher memorization under one abstraction tend to also exhibit higher memorization under others. However, the strength of this relationship vary: while it appears more structured in neural models, it becomes noisier and less pronounced in simpler architectures. This suggests that cross-abstraction memorization is a general tendency, but its strength depends on model capacity.}

{\paragraph{Effect of transformations.}
The analysis of segment-level transformations (Figures~\ref{fig:rq3_memorization_patterns_appendix}c,g,k,o) confirms that the type of perturbation influences memorization. In several models, \textit{shuffle} transformations tend to produce higher exposure values than \textit{substitute} or \textit{stationary} modes, indicating that stronger perturbations are harder for the model to generalize. However, the differences between transformation types are often moderate and can partially overlap, especially for models with mild memorization (e.g., \textit{LSTPM}). Moreover, absolute exposure values vary significantly across models, which makes direct visual comparison sensitive to scale. These results refine the main observation by showing that while transformation effects exist, they are not uniformly strong across architectures.}

{\paragraph{Temporal patterns.}
The temporal analysis (Figures~\ref{fig:rq3_memorization_patterns_appendix}d,h,l,p) reveals more heterogeneous behavior than suggested by the single-model analysis in the main paper. While some models (e.g., \textit{DeepMove}) show higher exposure during periods associated with behavioral transitions, these effects are generally less pronounced and not consistently observed across all architectures. In particular, simpler models such as \textit{Markov} exhibit relatively uniform exposure across time bins, reflecting their limited ability to capture temporal context. Other models, such as \textit{Graph-Flashback}, display higher variability without a clear temporal structure. Overall, temporal sensitivity appears to be model-dependent and weaker than suggested by the single-model case.}

\begin{figure*}[t]
\centering

\begin{subfigure}{0.49\linewidth}
    \centering
    \includegraphics[width=\linewidth]{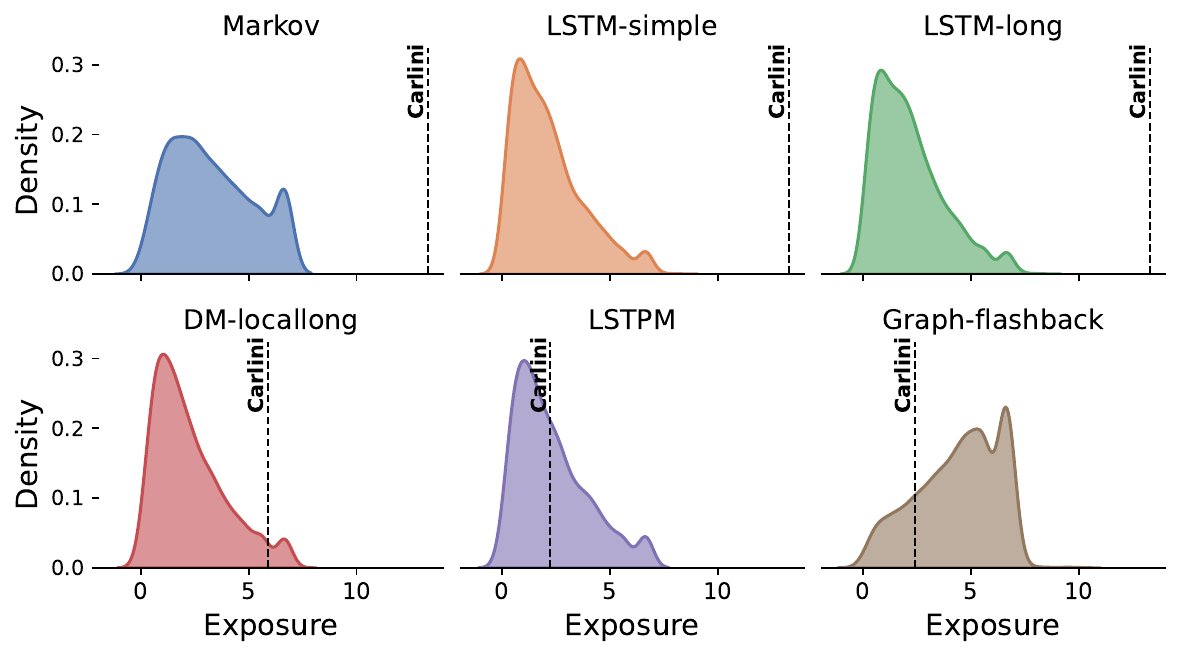}
    \caption{{Shanghai Kaggle}}
    \label{fig:enter-label-shanghai}
\end{subfigure}
\hfill
\begin{subfigure}{0.49\linewidth}
    \centering
    \includegraphics[width=\linewidth]{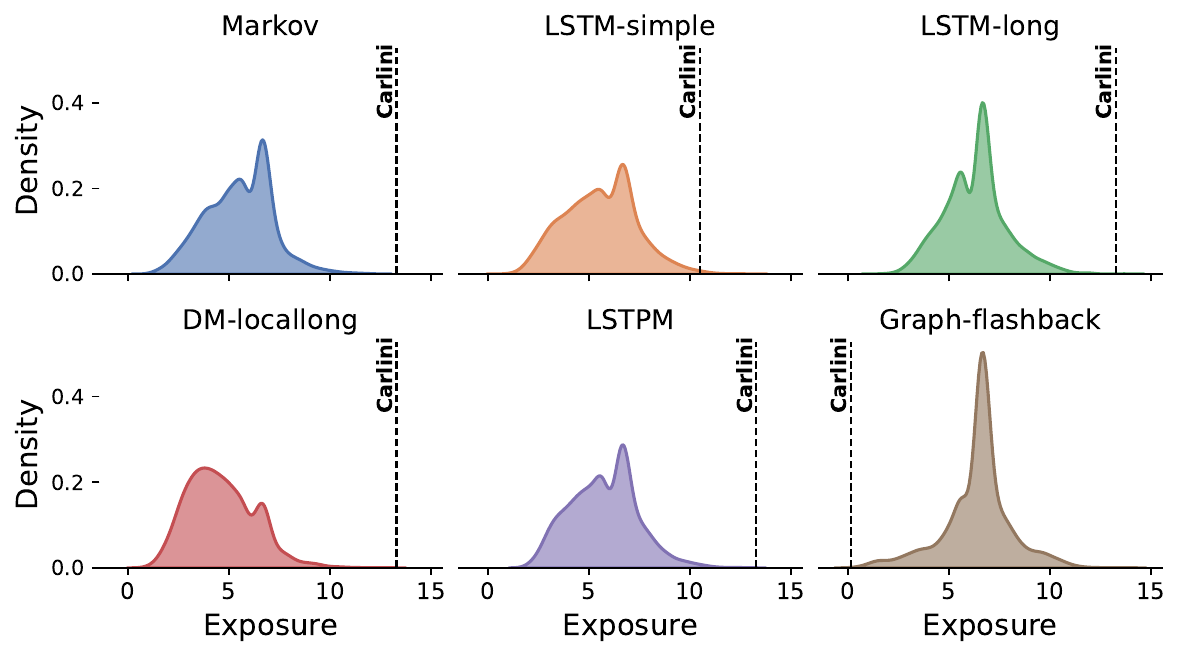}
    \caption{{YJMob100K}}
    \label{fig:enter-label-YJMob100K}
\end{subfigure}

\caption{
Exposure distributions across model architectures, with \citet{carlini2019secretsharer}'s synthetic–canary exposure markers shown for comparison.  
(a) {Shanghai Kaggle}.  
(b) {YJMob100K}.  
}
\label{fig:rq4_all_ridgeplots_appendix}
\end{figure*}

{\paragraph{Exposure distributions across models.}
Figure~\ref{fig:rq4_all_ridgeplots_appendix} extends the analysis of Figure~\ref{fig:enter-label} (main paper) to additional datasets. Consistent with the results in \textit{ShenzhenUrb}, exposure distributions remain highly model-dependent, with some architectures (e.g., \textit{Graph-Flashback}) exhibiting heavier tails and higher exposure values, while others (e.g., \textit{LSTPM}) show more compact distributions. However, the relative differences between models vary across datasets, and the alignment between user-level exposure distributions and the synthetic canary exposure markers remains inconsistent. These results further support the observation that memorization behavior depends on both the model architecture and the dataset characteristics.}

{\paragraph{Summary.}
Overall, the appendix results confirm the qualitative trends identified in the main paper—namely that memorization is stronger for structured, long-term patterns and weaker for localized segments—but also highlight important variability across models. In particular, (i) the magnitude of memorization differs substantially across architectures, (ii) the effect of transformations is present but not always pronounced, and (iii) temporal patterns are less consistent than suggested by the single-model analysis. These results emphasize that memorization behavior should be evaluated across multiple models to obtain a reliable assessment of privacy risks.}

{\subsection{Utility and Memorization Across Models}}
\label{appendix:rq4_additional}

\begin{figure*}[t]
\centering


\begin{subfigure}{0.16\textwidth}
    \centering
    \includegraphics[width=\linewidth]{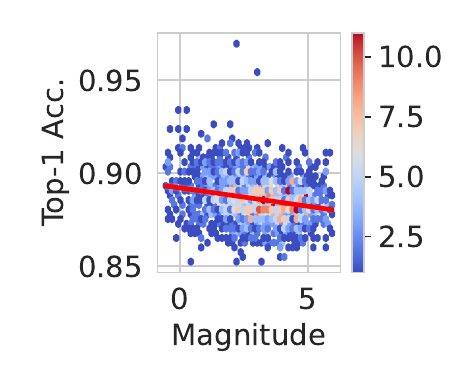}
    \caption{{Shenzhen Urban}}
\end{subfigure}
\hfill
\begin{subfigure}{0.16\textwidth}
    \centering
    \includegraphics[width=\linewidth]{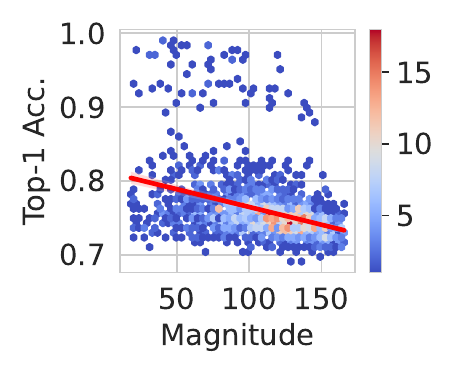}
    \caption{{Shanghai Kaggle}}
\end{subfigure}
\hfill
\begin{subfigure}{0.16\textwidth}
    \centering
    \includegraphics[width=\linewidth]{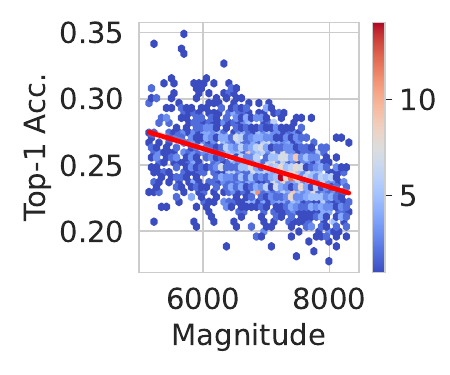}
    \caption{{YJMob100K}}
\end{subfigure}
\hfill
\begin{subfigure}{0.16\textwidth}
    \centering
    \includegraphics[width=\linewidth]{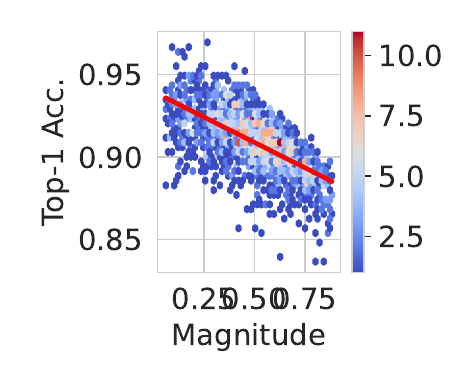}
    \caption{{Shenzhen Urban}}
\end{subfigure}
\hfill
\begin{subfigure}{0.16\textwidth}
    \centering
    \includegraphics[width=\linewidth]{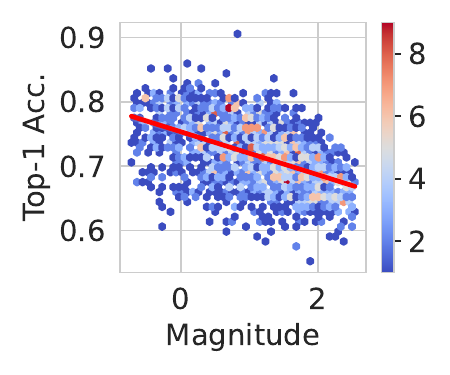}
    \caption{{Shanghai Kaggle}}
\end{subfigure}
\hfill
\begin{subfigure}{0.16\textwidth}
    \centering
    \includegraphics[width=\linewidth]{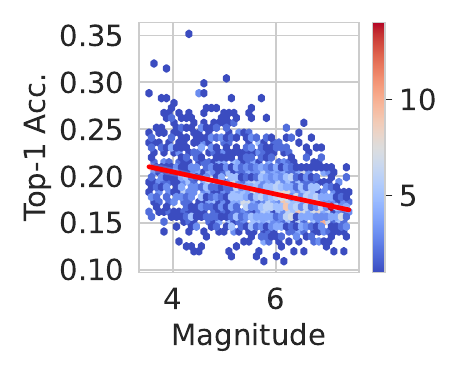}
    \caption{{YJMob100K}}
\end{subfigure}
\hfill
\begin{subfigure}{0.16\textwidth}
    \centering
    \includegraphics[width=\linewidth]{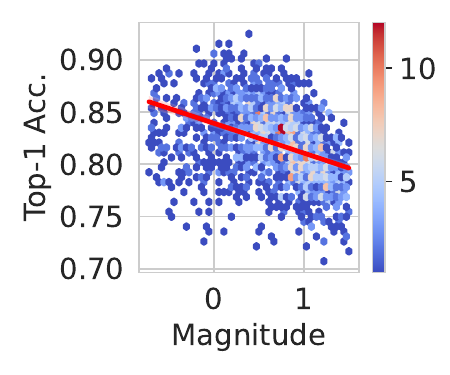}
    \caption{{Shenzhen Urban}}
\end{subfigure}
\hfill
\begin{subfigure}{0.16\textwidth}
    \centering
    \includegraphics[width=\linewidth]{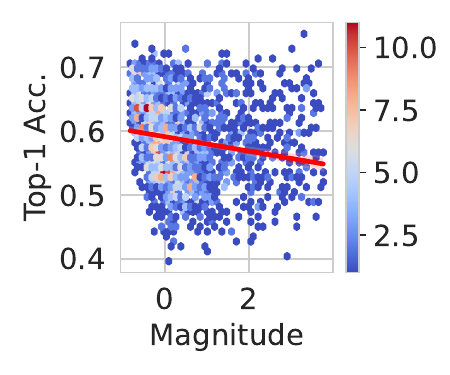}
    \caption{{Shanghai Kaggle}}
\end{subfigure}
\hfill
\begin{subfigure}{0.16\textwidth}
    \centering
    \includegraphics[width=\linewidth]{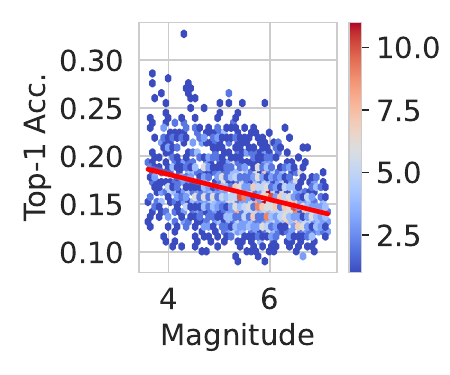}
    \caption{{YJMob100K}}
\end{subfigure}
\hfill
\begin{subfigure}{0.16\textwidth}
    \centering
    \includegraphics[width=\linewidth]{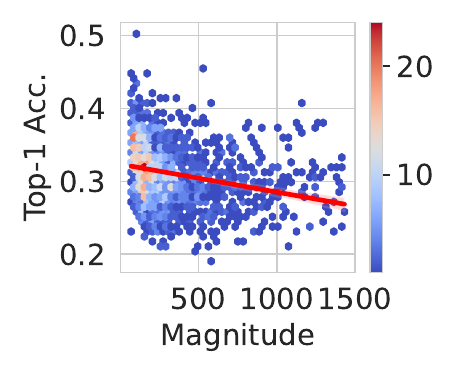}
    \caption{{Shenzhen Urban}}
\end{subfigure}
\hfill
\begin{subfigure}{0.16\textwidth}
    \centering
    \includegraphics[width=\linewidth]{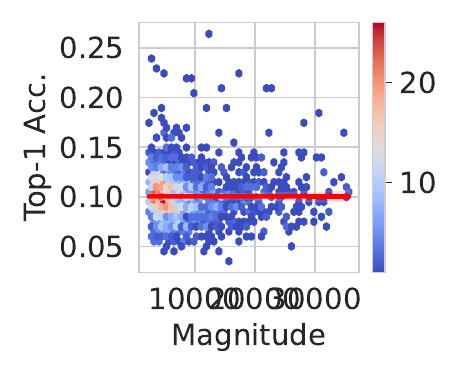}
    \caption{{Shanghai Kaggle}}
\end{subfigure}
\hfill
\begin{subfigure}{0.16\textwidth}
    \centering
    \includegraphics[width=\linewidth]{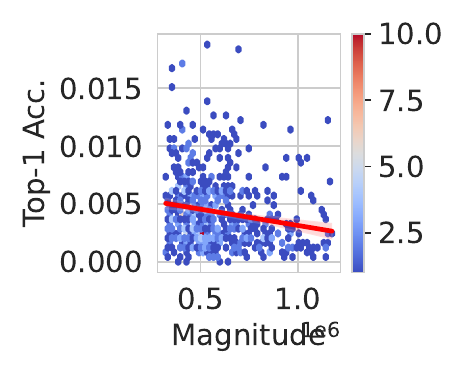}
    \caption{{YJMob100K}}
\end{subfigure}

\caption{
Memorization–utility relationship (magnitude vs.\ Top-1 Accuracy) across three datasets for the following models: (a)-(c) Markov, (d)-(f) DM-avglonguser, (g)-(i) LSTPM, and (j)--(l) Graph-flashback.
}
\label{fig:big_memo_utility_comparison}
\end{figure*}

{Figure~\ref{fig:big_memo_utility_comparison} extends the analysis of Figure~\ref{fig:rq4_memorization_vs_utility_hexbin} (main paper) by reporting the relationship between memorization (magnitude) and utility (Top-1 accuracy) across multiple model architectures.}

{Across models and datasets, we observe a consistent negative association between memorization and utility, where higher memorization tends to coincide with lower predictive accuracy. However, the strength and dispersion of this relationship vary across architectures. Simpler models exhibit more compact patterns, while neural models show broader variability.} 

{\paragraph{Summary.} These results support the main observation that the memorization–utility trade-off is present across settings, while highlighting that its magnitude depends on both the dataset and model.}

\begin{table*}[t]
\centering
\small
\setlength{\tabcolsep}{4pt}
\renewcommand{\arraystretch}{0.95}

\begin{tabular}{l l r r r r}
\toprule
\textbf{Memorization type} &
\textbf{Dataset} &
\textbf{Abstraction} &
\textbf{Clustering} &
\textbf{Ref. construction} &
\textbf{Total} \\
\midrule

Location
& ShanghaiTel   & $\sim$0.3  & $\sim$0.4 & $\sim$5.5 & $\sim$6 \\
& ShenzhenUrban & $\sim$4.6  & $\sim$3.4 & $\sim$12  & $\sim$20 \\
& YJMob100K     & $\sim$7.6  & $\sim$6.0 & $\sim$16  & $\sim$30 \\
\midrule

Anchor-pair
& ShanghaiTel   & $\sim$2.7  & $\sim$0.5 & $\sim$7   & $\sim$10 \\
& ShenzhenUrban & $\sim$37   & $\sim$1.7 & $\sim$21  & $\sim$60 \\
& YJMob100K     & $\sim$46   & $\sim$1.5 & $\sim$22  & $\sim$70 \\
\midrule

Segment-level
& ShanghaiTel   & -- & -- & $\sim$15 & $\sim$15 \\
& ShenzhenUrban & -- & -- & $\sim$15 & $\sim$15 \\
& YJMob100K     & -- & -- & $\sim$13 & $\sim$13 \\

\bottomrule
\end{tabular}

\caption{
Runtime (in minutes) for reference set construction across memorization abstractions and datasets.
The total runtime includes abstraction computation, clustering, medoid selection, reference set enrichment, and dataset generation.
Reference construction is performed only once per dataset and can subsequently be reused across models.
}
\label{tab:runtime_reference}
\end{table*}

\begin{table*}[t]
\centering
\small
\setlength{\tabcolsep}{3.5pt}
\renewcommand{\arraystretch}{0.95}

\begin{tabular}{l l r r r r r r r r r r r r}
\toprule

\textbf{Memorization type} &
\textbf{Dataset}
& \multicolumn{3}{c}{\textbf{Markov}}
& \multicolumn{3}{c}{\textbf{DeepMove}}
& \multicolumn{3}{c}{\textbf{LSTPM}}
& \multicolumn{3}{c}{\textbf{GraphFlashback}} \\

\cmidrule(lr){3-5}
\cmidrule(lr){6-8}
\cmidrule(lr){9-11}
\cmidrule(lr){12-14}

&
& \textbf{PPL} & \textbf{Met.} & \textbf{Tot.}
& \textbf{PPL} & \textbf{Met.} & \textbf{Tot.}
& \textbf{PPL} & \textbf{Met.} & \textbf{Tot.}
& \textbf{PPL} & \textbf{Met.} & \textbf{Tot.} \\

\midrule

\multirow{3}{*}{Location}
& ShanghaiTel
& $\sim$2.3 & $<0.1$ & $\sim$3
& $\sim$104 & $<0.1$ & $\sim$105
& $\sim$100 & $<0.1$ & $\sim$101
& $\sim$25 & $<0.1$ & $\sim$25 \\

& ShenzhenUrban
& $\sim$6 & $<0.1$ & $\sim$6
& $\sim$113 & $<0.1$ & $\sim$114
& $\sim$234 & $<0.1$ & $\sim$235
& $\sim$119 & $<0.1$ & $\sim$119 \\

& YJMob100K
& $\sim$5 & $<0.1$ & $\sim$7
& $\sim$123 & $<0.1$ & $\sim$124
& $\sim$314 & $<0.1$ & $\sim$316
& $\sim$53 & $<0.1$ & $\sim$53 \\

\bottomrule
\end{tabular}

\caption{
Runtime (in minutes) of location memorization assessment across datasets and models.
Runtime is dominated by perplexity computation (PPL), while memorization metric computation (Met.) requires only a few seconds across all datasets and models.
Experiments were conducted on a server equipped with an AMD EPYC 9124 16-core CPU, 1 TB RAM, and four NVIDIA A40 GPUs.
}
\label{tab:runtime_assessment}
\end{table*}

\begin{figure*}[htbp]
\centering

\begin{subfigure}{0.23\textwidth}
    \centering
    \includegraphics[width=\linewidth]{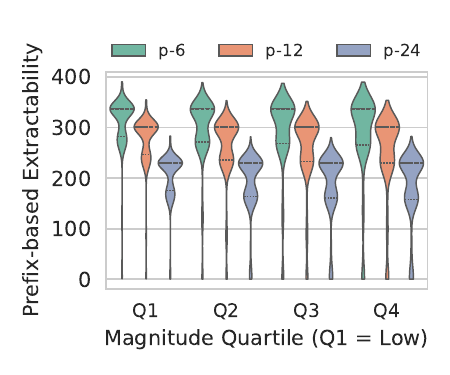}
    \caption{Prefix-based extractability}
    \label{fig:extractability_prefix_shenzhen}
\end{subfigure}
\hfill
\begin{subfigure}{0.23\textwidth}
    \centering
    \includegraphics[width=\linewidth]{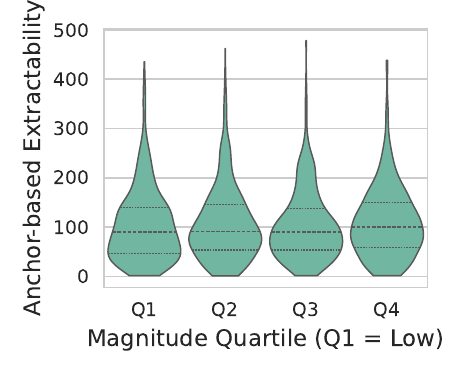}
    \caption{Anchor-based extractability}
    \label{fig:extractability_home_shenzhen}
\end{subfigure}
\hfill
\begin{subfigure}{0.23\textwidth}
    \centering
    \includegraphics[width=\linewidth]{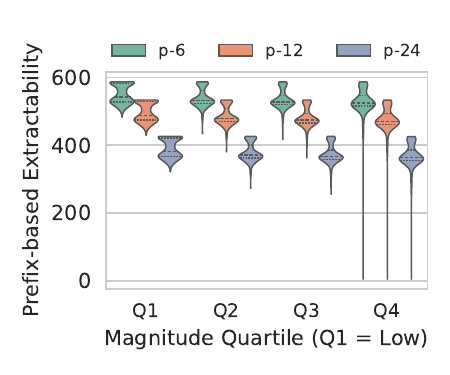}
    \caption{Prefix-based extractability}
    \label{fig:extractability_prefix_yjmob}
\end{subfigure}
\hfill
\begin{subfigure}{0.23\textwidth}
    \centering
    \includegraphics[width=\linewidth]{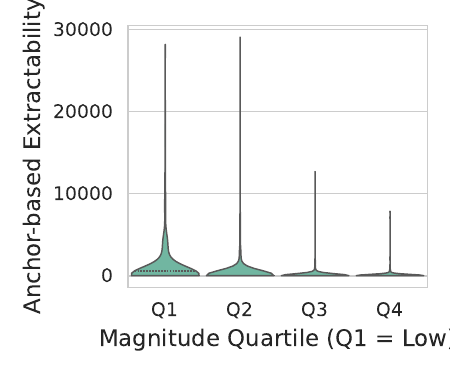}
    \caption{Anchor-based extractability}
    \label{fig:extractability_home_yjmob}
\end{subfigure}

\caption{
Extractability versus memorization signals for {LSTM-simple} across two datasets (a)-(b) {Shenzhen Urban} and (c)-(d) {YJMob100K}, under both prefix-based and anchor-based extraction attacks.
}
\label{fig:rq4_extractability_two_datasets}
\end{figure*}

\section{Extraction Risk Under Model Memorization}
\label{appendix:extractability_attacks}

To complement our analysis of memorization risks, we describe the design of two proxy extraction attacks—prefix-based and anchor-based—that simulate realistic adversarial inference scenarios. These proxies allow us to quantify how easily memorized trajectories can be recovered by a model-conditioned attacker. Each of them assumes access to a trained mobility prediction model which outputs a probability distribution over likely next locations. 
Both attacks rely on autoregressive beam search to explore the space of plausible trajectory continuations under partial user knowledge.

\subsection{Prefix-Based Extraction Attack}
\paragraph{Threat model.} The adversary has partial knowledge of a user's trajectory (e.g., the first day) and attempts to infer the remainder. This models scenarios where partial leaks expose only a fragment of user mobility.

\paragraph{Setup.} For each target trajectory, we define three levels of prior knowledge, using prefixes corresponding to 6, 12, and 24 hours. These prefixes are extracted from the first day of the reference trajectory (which spans 3 days total). The model is queried to predict the remaining hours. 

\paragraph{Decoding strategy.} We employ beam search \cite{attention-original} to decode the most likely suffixes. The search expands trajectory candidates autoregressively by maintaining the top-$k$ most probable sequences (with $k=5$) at each time step. Candidates are scored based on their cumulative log-likelihood as computed by the model.

\paragraph{Metric.} The vulnerability of a trajectory is evaluated by determining the rank of the ground-truth suffix within the model's likelihood distribution. We define the prefix extractability score as the log-scaled rank of the correct suffix:
\begin{equation}
\text{Extractability}_{\text{prefix}} = \log_{10}(\text{rank}_{\text{suffix}}).
\end{equation}
Lower scores indicate higher vulnerability (easier recovery). 

\subsection{Anchor-Based Extraction Attack}
\paragraph{Threat model.} The adversary knows one user's anchor location (i.e., home) and seeks to infer their other anchor (i.e., work). This captures risks associated with anchor-pair memorization enabling the inference of sensitive semantic locations (POIs).

\paragraph{Setup.} We simulate the attack by constructing a synthetic trajectory seeded with the true home location. The adversary assumes that their target stays at home from midnight until at least 6:00 and at work between 10:00 and 18:00. The ground-truth work location is inferred from the data and used for evaluation.

\paragraph{Decoding strategy.} Conditioning the model on the segment of the user's trajectory until 6:00, we perform beam search with size $k=5$ to find the $k$ most likely commute paths. 

\paragraph{Metric.} For each of the $k$ paths, we compute the rank of the true work anchor in the model's prediction distribution at each time step $t$ within work hours (10:00-18:00). The average of these values defines extractability:

\begin{equation}
\text{Extractability}_{\text{anchor}} = \frac{1}{k} \sum_{j=1}^k \frac{1}{T} \sum_{t=1}^{T} \text{rank}_{t,j}^{\text{work}},
\end{equation}
where $T=9\times2=18$ is the number of time-steps the workday takes (in our 30-minute interval case). Lower values indicate higher extractability.

\paragraph{Interpretation.} In both settings, memorized trajectories, identified by high magnitude scores, are consistently ranked more favorably, suggesting that memorization amplifies privacy risks in realistic adversarial settings.

{\subsection{Extractability Across Datasets}}
\label{appendix:extractability_results}

{We extend the analysis of extractability presented in the main paper (Figure~\ref{fig:rq4_memorization_vs_extractability}) to additional datasets in Figure~\ref{fig:rq4_extractability_two_datasets}.}

{\paragraph{Prefix-based extraction.}
In line with the main paper, trajectories with higher memorization magnitude generally tend to be easier to extract. However, the separation between magnitude quartiles is less pronounced in some datasets (e.g., Shenzhen Urban), where distributions across quartiles partially overlap. This indicates that while memorization contributes to extractability, it is not the sole determining factor, and its effect can vary depending on the dataset.}

{\paragraph{Anchor-based extraction.}
The relationship between anchor-pair memorization and extractability appears more variable across datasets. While a clearer trend is observed in the Shanghai Telecom dataset (main paper), Figure~\ref{fig:rq4_extractability_two_datasets} shows that this effect becomes much weaker in other datasets, with substantial overlap between magnitude quartiles and, in some cases, nearly indistinguishable distributions. }

{\paragraph{Discussion.}
Overall, these results suggest that the link between memorization and extractability is present but not uniformly strong. Memorization can facilitate extraction by making certain trajectories easier to recover, but its impact depends on both the dataset characteristics and the attack setting. This highlights that memorization is an important—but not exclusive—driver of privacy risk.}

\end{document}